# CPC-CMS: Cognitive Pairwise Comparison Classification Model Selection Framework for Document-level Sentiment Analysis

**Jianfei LI[1], Kevin Kam Fung YUEN[2*,1]**

**[1]Department of Computing, Faculty of Computer and Mathematical Sciences**

**The Hong Kong Polytechnic University, Hong Kong SAR, China**

**[2*]School of Science, Monash University Malaysia,**

**Jalan Lagoon Selatan, Malaysia**

[*]kevinkf.yuen@gmail.com, kevin.yuen@monash.edu

[*]https://orcid.org/0000-0003-1497-2575

**Abstract**

This study proposes the Cognitive Pairwise Comparison Classification Model Selection (CPC-CMS) framework for document-level sentiment analysis. The CPC, based on expert knowledge judgment, is used to calculate the weights of evaluation criteria, including accuracy, precision, recall, F1-score, specificity, Matthews Correlation Coefficient (MCC), Cohen's Kappa (Kappa), and efficiency. Naive Bayes (NB), Linear Support Vector Classification (LSVC), Random Forest, Logistic Regression, Extreme Gradient Boosting (XGBoost), Long Short-Term Memory (LSTM), and A Lite Bidirectional Encoder Representations from Transformers (ALBERT) are chosen as classification baseline models. A weighted decision matrix consisting of classification evaluation scores with respect to criteria weights is formed to select the best classification model for a classification problem. Three open social media datasets are used to demonstrate the feasibility of the proposed CPC-CMS. Based on our simulation, for evaluation results excluding the time factor, ALBERT performs best across all three datasets; if the time factor is included, no single model consistently outperforms the others. Through comparison, these conclusions are also supported by other aggregation and ranking methods, including Analytic Hierarchy Process (AHP), Technique for Order of Preference by Similarity to Ideal Solution (TOPSIS) and Multi-Objective Optimization by Ratio Analysis (MOORA), although aggregation values and ranks may vary. A sensitivity analysis using Spearman's Rank Correlation Test demonstrates the robustness of the proposed CPC-CMS framework. The CPC-CMS can be applied to other classification applications in various domains.

## 1. Introduction

Sentiment analysis is a subfield of natural language processing, aiming to extract and analyze sentiments and viewpoints from textual information to facilitate computational research on the attitudes and sentiments of entities such as topics, individuals, issues, and products [1]. The Internet and mobile phones have occupied a large part of people's lives, leading to an explosive growth in the number of platforms such as Twitter and online comments [2]. Automatically analyzing large volumes of comments quickly, efficiently, and at low cost to make corresponding decisions has generated great interest in various fields (e.g., e-commerce, healthcare, government departments, etc.) [3]. For example, in the field of healthcare, Baker et al. [4] reliably analyzed people's responses to colon cancer using the GRU model and made predictions for the disease. In the business sector, Rognone et al. [5] investigated the impact of news sentiment on the returns, trading volumes, and volatility of Bitcoin and traditional currencies.

Many sentiment analysis problems can be treated as classification problems, and they can be performed at three levels: the document level, the sentence level, and the aspect level [6]. For document level, a document containing all sentences is regarded as an independent object, expressing an overall opinion. For sentence level, a subjectivity label, either subjective or objective, is given to each sentence. Since objective sentences often do not convey any viewpoints, sentiment classification determines the sentiments (e.g., positive, neutral, or negative) of each subjective sentence in all sentences (a document). For the aspect level, a sentence requires a more detailed analysis. Aspect items and opinion items are extracted from the sentence to identify the sentiment of aspects. Sentence-level and aspect-level sentiment analysis are beyond the scope of this study. The focus of this study is to select the most appropriate classification model for the document-level sentiment classification problem.

Many classification models can be applied to sentiment analysis [7]. [8] reviewed different techniques on model evaluation, model selection and algorithm selection in machine learning. However, it appears that none of the techniques consider aggregating multiple evaluation criteria for the selection. A Multiple Criteria Decision Making (MCDM) approach may be a good way to address this issue. Furthermore, it appears that there is a lack of research including a human judgment method for classification model selection in document-level sentiment analysis. In this study, the Cognitive Pairwise Comparison Classification Model Selection (CPC-CMS) framework is proposed for document-level sentiment analysis. The CPC [9] [10] [11] is an ideal alternative of the Analytic Hierarchy Process (AHP) [12] [13] to determine the weights.

The term 'cognitive' within the CPC-CMS framework, derived from the Cognitive Network Process [10] [11], refers specifically to the human cognitive processes involved in comparative judgment. The Analytic Hierarchy Process (AHP) [12] [13] require experts to evaluate paired ratio scales, which can be cognitively taxing and prone to subjective inconsistencies. In contrast, CPC utilizes a paired interval scale, which aligns more naturally with human cognitive perception of differences. From a behavioral science perspective, it is fundamentally easier and less error-prone for a human decision-maker to cognitively assess the absolute difference between two criteria rather than their precise mathematical ratio. Thus, the 'cognitive' designation denotes this framework's alignment with human perceptual strengths, positioning it firmly as a genuinely human-centered multicriteria decision-support tool rather than a novel natural language processing algorithm.

The CPC [9] [10] [11], a human decision-making method, is applied to assign weights to the evaluation criteria of seven classification baseline models (Naive Bayes, Linear Support Vector Classification, Random

Forest, Logistic Regression, Extreme Gradient Boosting, Long Short-Term Memory, and A Lite Bidirectional Encoder Representations from Transformers (ALBERT)) across three datasets. A weighted decision matrix is then used to identify the best model.

The remainder of this paper is structured as follows.

- Section 2 provides an overview of the CPC-CMS framework.
- Section 3 describes the document-level sentiment analysis process, including document structure, preprocessing, and feature extraction.
- Section 4 introduces the classification baseline models.
- Section 5 presents the CPC method and provides an example illustrating the weight generation process using the Pairwise Opposite Matrix (POM) to construct the weighted decision matrix.
- Section 6 details the simulation environment, data manipulation, implementation, and results of the three case studies.
- Section 7 compares various MCDM methods, and Section 8 presents the sensitivity analysis of the CPC-CMS framework.
- Section 9 concludes the study and outlines directions for future research.

## 2. CPC-CMS Framework for document-level sentiment analysis

An overview of the Cognitive Pairwise Comparison Classification Model Selection (CPC-CMS) framework is presented in Figure 1. Each document in the dataset is labeled with the sentiment tags, such as "positive" and "negative", or more specific emotions like happiness, sadness, and anger [14]. Because these documents often originate from sources like social media, news reports, and product reviews, their quality can be unreliable due to a large amount of irrelevant information and noise. If these documents are not properly cleaned and transformed, the difficulty of model learning increases. Therefore, a series of preprocessing steps, such as text cleaning, tokenization, and stemming, are needed.

Furthermore, preprocessed text cannot be directly processed by machine learning models and must therefore be transformed into numerical features. Different feature extraction methods can capture document information at varying levels, and common techniques include Term Frequency–Inverse Document Frequency (TF-IDF), random initialization of word embeddings, and ALBERT.

This study will compare the performance and efficiency of seven classification baseline models using eight evaluation criteria (to avoid being misled by a single criterion). However, without determining the relative importance of each criterion, selecting the optimal model remains challenging. The CPC [9] [10] [11], a human-centered method, is used to determine the weights of the eight evaluation criteria. Finally, by aggregating the test results through a weighted selection matrix, the model achieving the highest overall score is identified as the optimal choice.

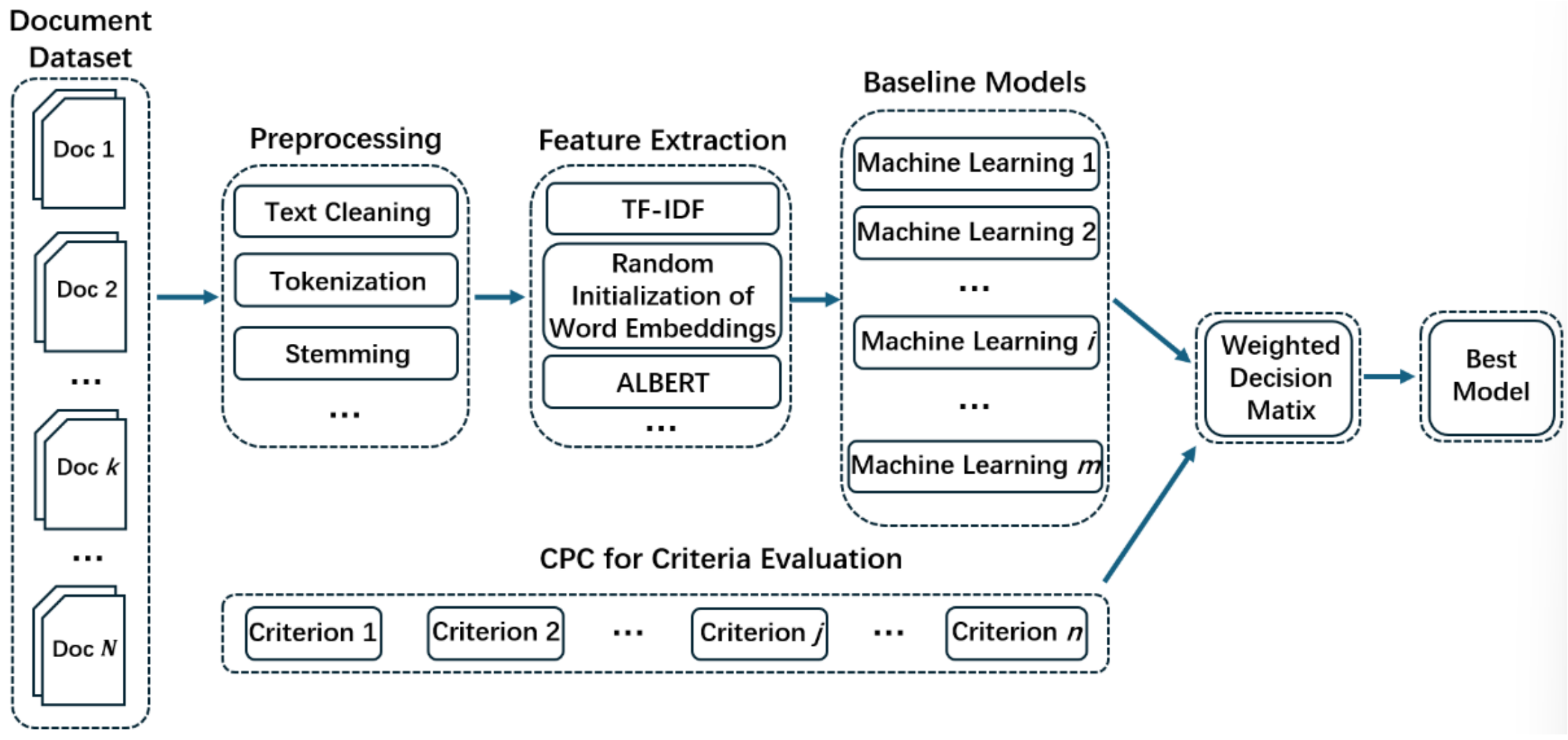

**Figure 1.** CPC-CMS Framework

## 3. Document-level Sentiment Analysis

### 3.1. Document Structure

The purpose of document-level sentiment analysis is to determine the overall opinion conveyed by an entire document. In document-level sentiment classification, handling a document containing sentences with mixed sentiments is a challenging task. However, the focus of this study is not on resolving sentiment contradictions within a document; interested readers are referred to [15] [16]. Although document-level sentiment analysis cannot achieve highly specific or fine-grained classification, it still has valuable application scenarios, such as measuring overall user satisfaction with a product, and many studies are based on this approach. For example, Zhai et al. [17] introduced an autoencoder loss function that incorporates sentiment information during the learning stage to obtain better document vectors and thereby achieve sentiment classification. Xu et al. [18] proposed a Cached Long Short-Term Memory (CLSTM) neural network to capture the overall semantic information in long texts.

Document-level structure treats an entire document—comprising all its constituent sentences—as a single entity and assigns it an overall sentiment label. Let a document $D$ be a sequence of sentences $\{s_1, s_2, \ldots, s_\alpha\}$. The goal is to predict the corresponding sentiment category $y$ (e.g., positive, neutral, or negative). A classification model $M$ is to map $D$ to $y$:

$$y = M(D) = M(\{s_1, s_2, \ldots, s_\alpha\}) \tag{1}$$

**Table 1.** Example of one document

| **Document** | My life is chaos. There is no solution. Fear of the uncertain. Restless direction. |
|---|---|
| **Number of sentences** | 4 |
| **Sentiment** | Anxiety |

To illustrate the document structure used in this study, Table 1 a sample document from the dataset [19] analyzed in Case 1 of Section 6.2. The document consists of four sentences: $s_1$ = "My life is chaos.", $s_2$ = "There is no solution.", $s_3$ = "Fear of the uncertain.", and $s_4$ = "Restless direction.". The sentiment label, "Anxiety", represents the overall sentiment of this document. Thus, $y = M(D) = M(\{s_1, s_2, s_3, s_4\}) =$ Anxiety.

### 3.2. Preprocessing

Data preprocessing is the process of converting raw data into a standardized and analyzable format. Documents are sometimes redundant and may have missing values for certain features, requiring the deletion of words and other components (such as special characters) that do not contain relevant information. There are several steps to clean all documents:

- Remove the RTs, @, #, and the links from the sentences.
- Convert the words to lowercase.
- Remove all non-alphabetic characters, including numbers and punctuation marks (optional).
- Eliminate redundant spaces.
- Decode HTML into plain text.

**Table 2.** Example of a cleaned document

| | |
|---|---|
| **The original document** | My life is chaos. There is no solution. Fear of the uncertain. Restless direction. |
| **The cleaned document 1** | my life is chaos there is no solution fear of the uncertain restless direction |
| **The cleaned document 2** | my life is chaos. there is no solution. fear of the uncertain. restless direction. |

Table 2 illustrates an example of cleaning a document. Every letter in the cleaned document is converted to lowercase (e.g., “My" is converted to "my", "There" is converted to "there", etc.). If the step of removing all non-alphabetic characters is performed, the result is presented in Cleaned Document 1, without any punctuation marks. If the punctuation is retained, the result is presented in Cleaned Document 2, which is used for ALBERT.

**Table 3.** Example of the three forms of the token set

| **Document 1** | my life is chaos there is no solution fear of the uncertain restless direction | **Classification baseline models** |
|---|---|---|
| **Token set 1** | ['my', 'life', 'is', 'chaos', 'there', 'is', 'no', 'solution', 'fear', 'of', 'the', 'uncertain', 'restless', 'direction'] | Naive Bayes, Random Forest, LSVC, Logistic Regression, and XGBoost |
| **Token set 2** | [6, 39, 11, 3716, 70, 11, 52, 1434, 349, 9, 5, 4450, 1178, 2123] | LSTM |
| **Document 2** | my life is chaos. there is no solution. fear of the uncertain. restless direction. | |
| **Token set 3** | ['[CLS]', '▁my', '▁life', '▁is', '▁chaos', '.', '▁there', '▁is', '▁no', '▁solution', '.', '▁fear', '▁of', '▁the', '▁uncertain', '.', '▁restless', '▁direction', '.', '[SEP]'] | ALBERT |

As the cleaned documents are further tokenized, Table 3 shows examples of different tokenization formats used by different classification baseline models. For Naive Bayes, Random Forest, LSVC, Logistic Regression, and XGBoost, the document is segmented by the spaces and punctuation marks, and converted into lists of strings with one token representing one word (e.g., Token set 1 in Table 3). For LSTM, the document is also segmented by the spaces and punctuation marks, and the segmentation results are converted into numbers based on the vocabulary list (e.g., for Token set 2 in Table 3, "my" is mapped to 6 and "life" is mapped to 39). ALBERT uses a pretrained tokenizer to tokenize words (e.g., Token set 3 in Table 3). [CLS] indicates the beginning of the document, and [SEP] indicates the end of the document or a separator between two documents.

**Table 4.** Example of Stemming

| | |
|---|---|
| **Document** | my life is chaos there is no solution fear of the uncertain restless direction |
| **Stemming** | my life is chao there is no solut fear of the uncertain restless direct |

Stemming is the process of simplifying words into their base forms. In Table 4, "chaos" becomes "chao" and "solution" becomes "solut". Stemming can reduce feature dimensionality and sparsity but may weaken the language structures of LSTM and ALBERT. Thus, stemming applied only to the other classification baseline models in this study.

### 3.3. Feature Extraction

Different classification baseline models use different feature extraction methods. In this study, LSTM used the random initialization of word embeddings; ALBERT used its pre-trained word embeddings. The rest of the models used Term Frequency–Inverse Document Frequency (TF-IDF).

#### 3.3.1. Term Frequency–Inverse Document Frequency

TF-IDF (Term Frequency–Inverse Document Frequency) [20] is a statistical metric used to reflect the importance of a word to a document in a corpus. TF-IDF counts and calculates the frequency of a word occurring in the target document and that of its occurrence in other documents contained in the corpus. A higher importance score results from a higher occurrence frequency in the target document and a lower occurrence frequency in other documents of the corpus.

TF-IDF is used to convert the documents into a weighted bag-of-words vector and calculate the weight of each word. There are various calculation formulas for Term Frequency (TF), but the most intuitive definition of TF is the number of times of a word $t$ appears in the document $D$:

$$TF(t, D) = count(t \; in \; D) \quad (2)$$

Suppose $N$ is the total number of documents in the corpus, and the Document Frequency $DF(t)$ represents the number of documents containing the word $t$. Inverse Document Frequency ($IDF$) is used to measure the global importance of the word $t$.

$$IDF(t) = \ln\left(\frac{1+N}{1+DF(t)}\right)+1 \quad (3)$$

Combining the above two formulas yields $TF\text{-}IDF$.

$$TF\text{-}IDF(t,D) = TF(t,D) \times IDF(t) \quad (4)$$

Finally, L2 normalization is performed on the resulting $TF\text{-}IDF$ vector.

$$Norm\big(TF\text{-}IDF(t,D)\big) = \frac{TF\text{-}IDF(t,D)}{\sqrt{\sum_t TF\text{-}IDF(t,D)^2}} \quad (5)$$

**Table 5.** Example of different stages from Token to normalized TF-IDF

| Token | TF | DF | IDF | TF-IDF | Normalized TF-IDF |
|---|---|---|---|---|---|
| chao | 1 | 54 | 7.6415 | 7.6415 | 0.3511 |
| direct | 1 | 241 | 6.1599 | 6.1599 | 0.2830 |
| fear | 1 | 1374 | 4.4227 | 4.4227 | 0.2032 |
| fear of | 1 | 485 | 5.4627 | 5.4627 | 0.2510 |
| is | 2 | 21184 | 1.6878 | 3.3756 | 0.1551 |
| is no | 1 | 961 | 4.7799 | 4.7799 | 0.2196 |
| life | 1 | 9849 | 2.4536 | 2.4536 | 0.1127 |
| life is | 1 | 1692 | 4.2146 | 4.2146 | 0.1936 |
| my | 1 | 22967 | 1.607 | 1.607 | 0.0738 |
| my life | 1 | 5381 | 3.0581 | 3.0581 | 0.1405 |
| no | 1 | 8962 | 2.548 | 2.548 | 0.1171 |
| no solut | 1 | 13 | 9.0098 | 9.0098 | 0.4140 |
| of | 1 | 21660 | 1.6656 | 1.6656 | 0.0765 |
| of the | 1 | 4714 | 3.1904 | 3.1904 | 0.1466 |
| restless | 1 | 310 | 5.9091 | 5.9091 | 0.2715 |
| solut | 1 | 307 | 5.9188 | 5.9188 | 0.2719 |
| the | 1 | 25515 | 1.5018 | 1.5018 | 0.0690 |
| there | 1 | 7464 | 2.7309 | 2.7309 | 0.1255 |
| there is | 1 | 2729 | 3.7368 | 3.7368 | 0.1717 |
| uncertain | 1 | 38 | 7.9853 | 7.9853 | 0.3669 |

The token values and their normalized TF-IDF scores at different stages are presented in Table 5. For example, to illustrate this, the calculation process for the normalized TF-IDF of "chao" is shown below.

In the "Document" row of Table 4, since "chao" appears once in the document, the TF of "chao" is 1. There are a total of 42144 documents, and "chao" appears 54 times in other documents. The IDF of "chao" can be calculated using Eq. 3.

$$IDF(\text{"chao"}, D) = \ln\left(\frac{1+42144}{1+54}\right)+1 \approx 7.6415$$

TF-IDF of "chao" is obtained using Eq. 4.

$$TF\text{-}IDF(\text{"chao"}, D) = 1 \times 7.6415 = 7.6415$$

Finally, the normalization of the TF-IDF of "chao" is calculated using Eq. 5.

$$Norm(TF\text{-}IDF(\text{"chao"}, D)) = \frac{7.6415}{\sqrt{(7.6415)^2 + (6.1599)^2 + \cdots + (7.9853)^2}} \approx 0.3511$$

The remaining tokens are calculated using the same method demonstrated above, and the final results are shown in Table 5.

### 3.3.2. Random Initialization of Word Embeddings

Word embedding [21] is a technique used for language modelling and feature learning. Word embedding maps a discrete word to a continuous vector, and each dimension represents a word feature. Therefore, similar vector representations indicate similar meanings for the corresponding words. In this study, LSTM used the random initialization of word embeddings, which is a process where word vectors are initialized with random values at the beginning of model training, and then iteratively updated. For example, if the input dimension of the LSTM is 64, the "6" in "Token set 2" of Table 3 is mapped to a 64-dimensional vector [0.00746527, 0.01807993, -0.04794213, 0.03967083, -0.01264023, ⋯].

### 3.3.3. ALBERT pre-trained word embeddings

Bidirectional Encoder Representations from Transformers (BERT) is a deep bidirectional language model based on Transformers, supplemented by word and position encoding [22]. A Lite BERT (ALBERT) [23] is a lightweight version of BERT. ALBERT can achieve deep bidirectional representations in an unlabeled document since its context word embeddings include inter-sentence relationships and context (formed after understanding the entire annotation based on the semantic meanings of the retained words). ALBERT uses the sum of three embeddings of the token $x_i$: the word embeddings denoted by $E_{token}(x_i)$, the position embeddings denoted by $E_{pos}(x_i)$, and the sentence embeddings, $E_{seg}(x_i)$, to distinguish different documents.

$$E(x_i) = E_{token}(x_i) + E_{pos}(x_i) + E_{seg}(x_i) \qquad (6)$$

For example, if $x = \{x_i\}$ is Token set 3 in Table 3, $\{E_{token}(x_i)\}$ is shown below.

$$[2, 51, 201, 25, 8205, 9, 80, 25, 90, 4295, 9, 1719, 16, 14, 8373, 9, 15487, 1400, 9, 3]$$

ALBERT usually adds an attention mask to determine which positions are real words and which are padding. The number of tokens in the above Token IDs determines the number of "1"s in the attention mask (no padding); thus, the attention mask $\{E_{pos}(x_i)\}$ is shown below:

$$[1, 1, 1, 1, 1, 1, 1, 1, 1, 1, 1, 1, 1, 1, 1, 1, 1, 1, 1, 1]$$

If the sentence is padded to 128 tokens, the attention mask adds one hundred and eight "0" after the first twenty "1"s, as shown below:

$$[1, 1, 1, 1, 1, 1, 1, 1, 1, 1, 1, 1, 1, 1, 1, 1, 1, 1, 1, 1, 0, 0, \ldots, 0, \ldots, 0]$$

To distinguish between two documents, the Segment ID of the first document will be set to 0, and that of the second document to 1. If the number of documents mentioned above is only one, the Segment IDs are omitted or all set to 0, and the length is consistent with the number of tokens in the Token IDs. Suppose the tokens of the example document are from the first one; the Segment IDs $\{E_{seg}(x_i)\}$ are shown below:

$$[0, 0, 0, 0, 0, 0, 0, 0, 0, 0, 0, 0, 0, 0, 0, 0, 0, 0, 0, 0, 0]$$

By adding the Token IDs, attention mask, and Segment IDs together, the final input representation is obtained.

## 4. Classification Baseline Models

### 4.1. Naïve Bayes classifier

Given some evidence, the probability of a hypothesis can be calculated. For a token sequence $x$ and the $j$-th sentiment category $y_j$, Bayes' theorem can be used:

$$P(y = y_j|x) = \frac{P(x|y = y_j)P(y = y_j)}{P(x)} \tag{7}$$

Bernoulli Naive Bayes (BernoulliNB) [24] is a variant of the Naive Bayes algorithm. In BernoulliNB, each feature is binary (0 or 1), and due to the conditional independence assumption of Naive Bayes, features are independent of each other under a given category. Therefore, the likelihood of category $y_j$ is:

$$P(x|y = y_j) = \prod_{i=1}^{\mu} [P(x_i = 1|y = y_j)]^{x_i} \cdot [1 - P(x_i = 1|y = y_j)]^{1-x_i} \tag{8}$$

The predicted classification result, $\hat{y}$, is calculated as follows:

$$\hat{y} = \underbrace{argmax}_{y} \left[ P(y = y_j) \cdot \prod_{i=1}^{\mu} [P(x_i|y = y_j)] \right] \tag{9}$$

### 4.2. Random Forest

Figure 2 shows the specific process of Random Forest [25] algorithm. The Random Forest includes the Bagging algorithm and random feature selection. The Bagging algorithm generates multiple subsets from the original dataset through sampling with replacement (bootstrap), and a decision tree is generated from each sub-data set.

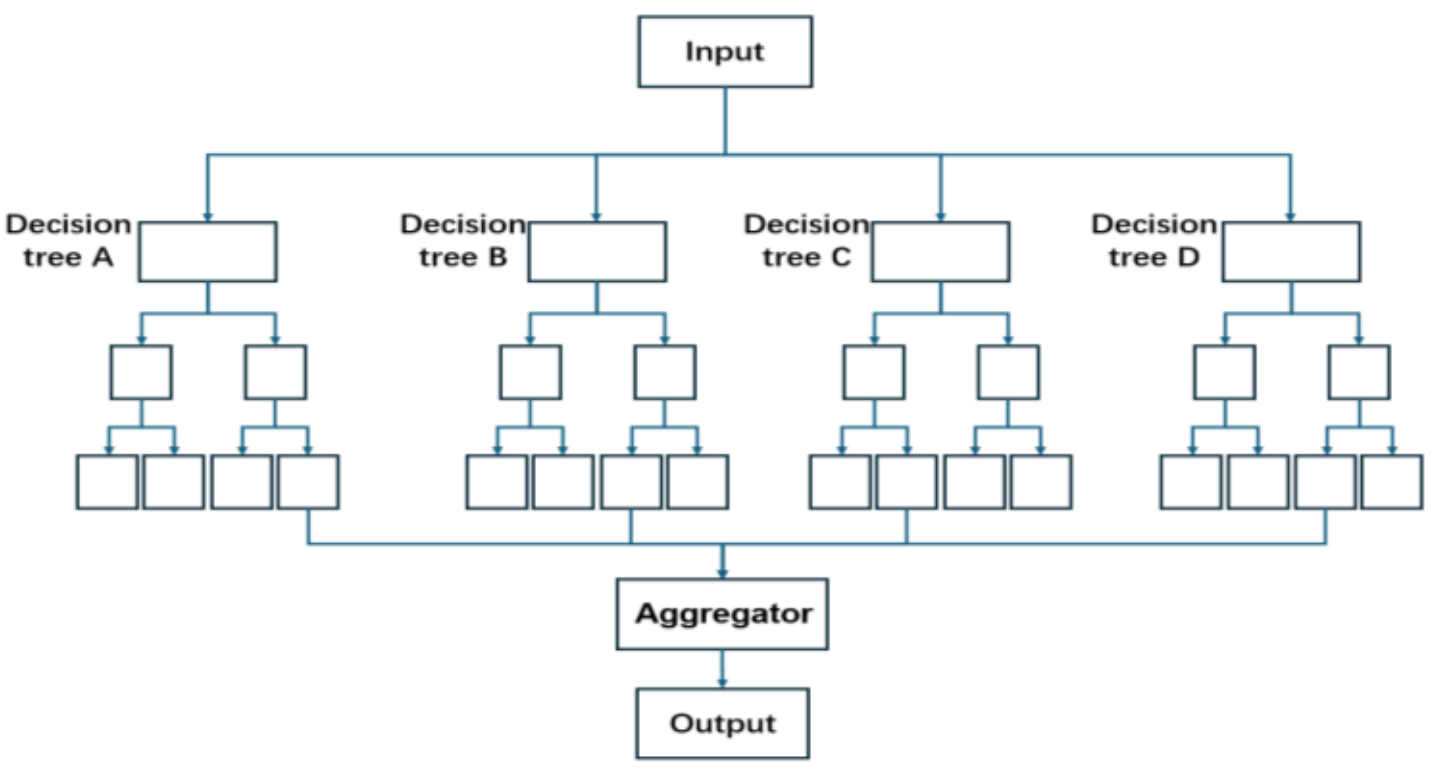

**Figure 2.** Example of Random Forest

Random feature selection randomly selects some features for comparison at each node splitting point. Let $N_l$ be the number of category samples in the leaf node where the sample falls, and $N_m$ be the total number of all samples in this leaf node. For the $i$-th tree, its prediction function $f_i(x)$ is shown below:

$$f_i(x) = \underbrace{argmax}_{y} \left( \frac{N_l^{(i)}}{N_m^{(i)}} \right) \tag{10}$$

A single decision tree is very sensitive to noise in the training set, but the Bagging algorithm reduces the correlation among the trained decision trees, which effectively alleviates this problem. The result is obtained through voting or averaging. Random Forest votes or averages the prediction results of all the trees ($A$ indicates selecting the majority category):

$$\hat{y}(x) = A(f_1(x), f_2(x), \dots, f_\eta(x)) \tag{11}$$

### 4.3. Logistic regression

Logistic regression is a linear algorithm that applies maximum likelihood estimation and is often used in binary classification problems [26]. Suppose $\theta^T = [\theta_1, \theta_2, \dots, \theta_\mu]^T$ are the weights of logistic regression and $\bar{\theta}$ is the bias of logistic regression. Let $z$ be the prediction value of the logistic regression:

$$z = \theta_0 + \theta_1 x_1 + \theta_2 x_2 + \cdots + \theta_\mu x_\mu = \bar{\theta} + \theta^T x \tag{12}$$

Logistic regression uses the sigmoid function $\sigma$ to model the prediction value and map the results to the interval [0, 1] as the probability of a sample belonging to a certain category:

$$\sigma(z) = \frac{1}{1 + e^{-z}} \tag{13}$$

For $\pi$ categories, the probability of the predicted category $y_i$ is calculated as shown below:

$$P(y = y_i | x) = \frac{e^{\theta_{y_i}^T x + \bar{\theta}_i}}{\sum_{i=1}^{\pi} e^{\theta_i^T x + \bar{\theta}_i}} \tag{14}$$

### **4.4.** Linear Support Vector Classification

LSVC (Linear Support Vector Classification) [27] is a specific implementation of a linear Support Vector Machine. The advantage of LSVC is that it can be applied to larger-scale datasets. Usually, the optimization algorithm based on Coordinate Descent is used to minimize the objective function. Suppose $\vartheta^T = [\vartheta_1, \vartheta_2, \dots, \vartheta_\mu]^T$ are the weights of LSVC and $\vartheta_0$ is the bias of LSVC. For a token sequence $x$, the linear score of the predicted category, $ls$, can be calculated as follows.

$$ls(x) = min\left(\frac{1}{\lambda}\sum_{i=1}^{\pi} \ell(y_i, \vartheta^T x + \vartheta_0) + \lambda\|\vartheta_s\|_1\right) \tag{15}$$

The Squared Hinge Loss is used as the loss function $\ell(y_i, \theta^T x_i + \theta_0)$. This loss function imposes a square penalty on samples with incorrect classification or insufficient classification confidence. The penalty for incorrect samples can help the model to separate categories more strictly. The calculation of the loss function is as follows.

$$\ell(y_i, \vartheta^T x + \theta_0) = max(0, 1 - y_i(\theta^T x + \theta_0))^2 \tag{16}$$

$\|\vartheta_s\|_1$ represents the L1 regularization term for some weights of the model. To achieve feature selection (sparsity), the L1 regularization term causes some weights in the model to become 0. $\lambda$ is the coefficient of the regularization term. The smaller $\lambda$ is, the weaker the regularization is. The formula of the L1 regularization term is as follows.

$$\lambda\|\vartheta^T\|_1 = \lambda\sum_{i=1}^{\mu}|\vartheta_i| \tag{17}$$

LSVC itself does not output a probability but a real score. However, in multi-classification tasks, these scores can be used for comparison. For $\pi$ categories, the predicted category $\hat{y}$ is as follows.

$$\hat{y} = \arg max_{i=1,\ldots,\pi}(\vartheta^T x + \vartheta_0) \tag{18}$$

## 4.5. Extreme Gradient Boosting

XGBoost (Extreme Gradient Boosting) [28] is an ensemble learning algorithm based on Gradient Boosting Decision Tree (GBDT), which is used for classification and regression tasks. The core idea is to gradually improve the model's performance by iteratively training multiple weak learners (usually decision trees), with each tree fitting the residuals of the previous tree. The final predicted value of XGBoost is the weighted sum of the outputs of all trees. Suppose $\hat{y}_i^r$ is the predicted output of the $i$-th sample $\bar{x}_i$ in the $r$-th round, and $f_j(\bar{x}_i)$ is the output of the $j$-th tree (i.e., the score of the leaf nodes). $\hat{y}_i^r$ can be calculated as follows:

$$\hat{y}_i^r = \sum_{j=1}^{r}(f_j(\bar{x}_i)) \tag{19}$$

The objective function of XGBoost consists of a loss function and a regularization term, which are used to measure the prediction error and complexity of the model. Suppose $\bar{y}_i$ is the true category of the $i$-th sample, $L$ is the loss function, and $\Omega$ is the complexity penalty term to prevent overfitting. The objective function can be calculated as follows:

$$Obj^{(r)} = \sum_{i=1}^{\pi} L\left(\bar{y}_i, \hat{y}_i^r\right) + \sum_{j=1}^{r}\Omega(f_j) \tag{20}$$

Suppose $y_k$ is the true category, and $\hat{y}_k$ is the predicted category. The loss function $L$ is calculated as follows:

$$L(y_k, \hat{y}_k) = -\sum_{k=1}^{\pi} y_k \cdot \log\left(\frac{1}{1+e^{-\hat{y}_k}}\right) \tag{21}$$

Suppose $N_{leaf}$ is the number of leaf nodes, and $\omega_k$ is the output weight of the $k$-th leaf node. $\gamma$ is a penalty term that controls the number of leaf nodes, and $\varepsilon$ is the L2 regularization term that controls the weight size of the leaves. The complexity penalty term $\Omega$ is calculated as follows:

$$\Omega(f_k) = \gamma N_{leaf} + \frac{1}{2}\varepsilon \sum_{k=1}^{N_{leaf}} \omega_k \tag{22}$$

### 4.6. Long Short-Term Memory

LSTM (Long Short-Term Memory) [29] is a special type of recurrent neural network (RNN), which is especially suitable for processing sequential data, such as text, time series, etc. LSTM introduces three gates (input gate, forget gate, and output gate) and a cell state by introduction, which allow it to better capture long-term dependencies efficiently, thus excelling in many natural language processing tasks.

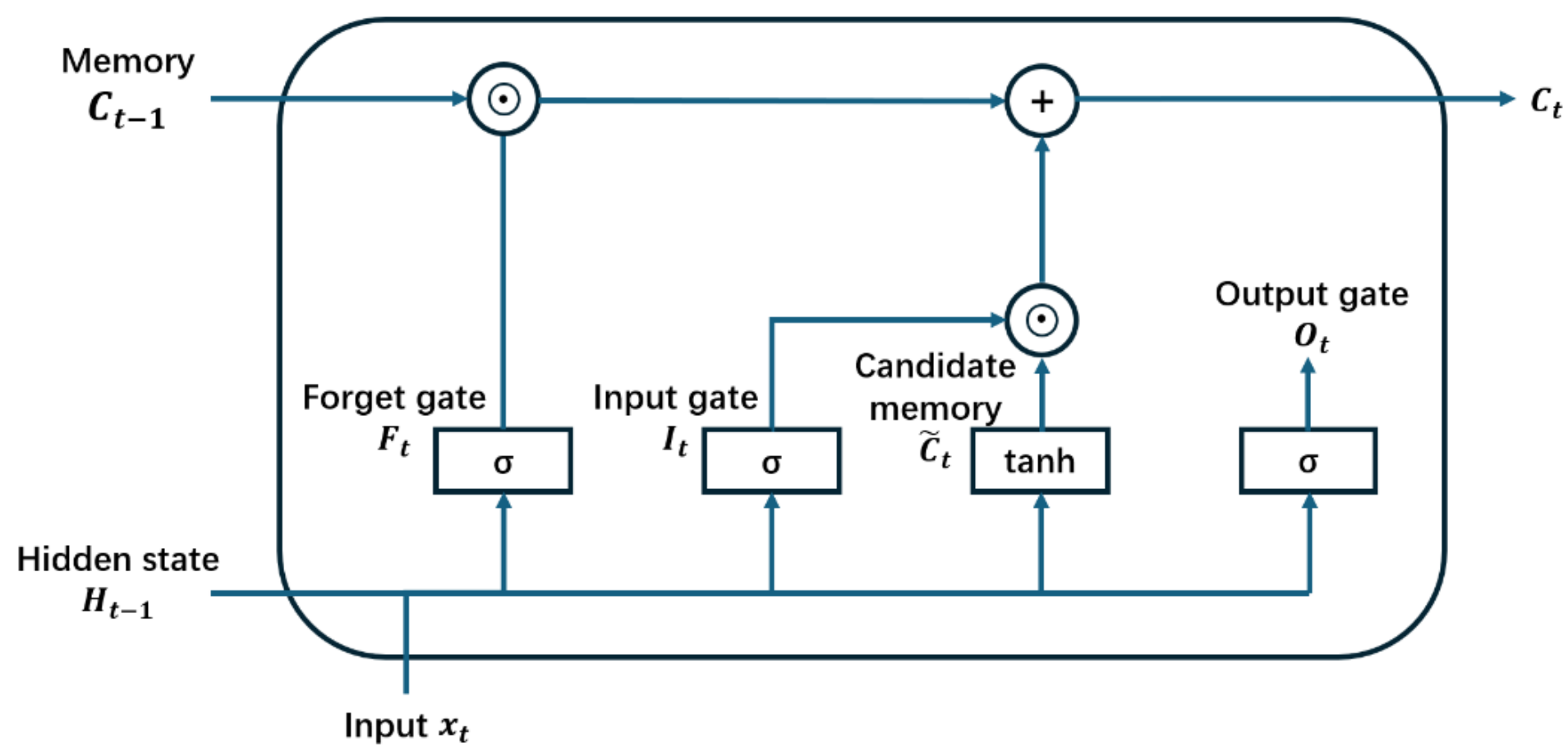

**Figure 3.** Example of LSTM

In Figure 3, $x_t$ is the input information and $H_{t-1}$ is the hidden state. The input gate, $I_t$, adds the information that needs to be retained from the previous state and the information that needs to be remembered now to get the candidate memory state $\tilde{C}_t$. Let $W_i$ be the weight of $I_t$, $b_i$ be the bias of $I_t$, $W_c$ be the weight of $\tilde{C}_t$, and $b_c$ be the bias of $\tilde{C}_t$. Here, tanh is the hyperbolic tangent function.

$$I_t = \sigma(W_i \cdot [H_{t-1}, x_t] + b_i) \tag{23}$$

$$\tilde{C}_t = \tanh(W_c \cdot [H_{t-1}, x_t] + b_c) \tag{24}$$

The forget gate, $F_t$, discards a portion of the previous memory from the memory cell. Suppose $W_f$ is the weight of $F_t$, and $b_f$ is the bias of $F_t$. $F_t$ is calculated as follows.

$$F_t = \sigma\left(W_f \cdot [H_{t-1}, x_t] + b_f\right) \tag{25}$$

The notation $\odot$ represents element-wise multiplication. The memory cell $C_t$ updates its state as follows.

$$C_t = F_t \odot C_{t-1} + I_t \odot \tilde{C}_t \tag{26}$$

Through this operation and the sigmoid function $\sigma$, we get a vector with values between 0 and 1, where 0 corresponds to parts of the previous memory to be forgotten, and 1 corresponds to parts to be retained. The output gate $O_t$ determines what information in the memory unit will be output. Then, the current hidden state $H_t$ is calculated. Let $W_o$ be the weight of $O_t$, and $b_o$ be the bias of $O_t$. The formulas are as follows:

$$O_t = \sigma(W_o \cdot [H_{t-1}, x_t] + b_o) \tag{27}$$

$$H_t = O_t \odot \tanh(C_t) \tag{28}$$

### 4.7. A Lite Bidirectional Encoder Representations from Transformers

The Transformer [30] is a deep learning model based on a self-attention mechanism that enables the model to consider all locations in the input sequence at the same time, rather than processing step by step like recurrent neural networks (RNNs) or convolutional neural networks (CNNs).

As shown in Figure 4, A Lite Bidirectional Encoder Representations from Transformers (ALBERT) [23] is a Transformer architecture based on the encoder structure, and its core components include Multi-Head Self-Attention, Feed-Forward Network, Layer Norm, and Residual Connections. The input text is first tokenized into a token sequence $x = [x_1, x_2, \dots, x_i, \dots, x_n]$. $P_{token}$ represents the low-dimensional word embeddings, and a feed-forward projection, $U$, is used to map the low-dimensional word embeddings back to the high-dimensional word embeddings. $E_{pos}$ represents the position embeddings, and $E_{seg}$ represents the sentence embeddings. Each token is then converted into an embedding vector:

$$E(x_i) = P_{token}(x_i) \cdot U + E_{pos}(x_i) + E_{seg}(x_i) \tag{29}$$

The input representation of the entire sequence is:

$$E(x) = [E(x_1), E(x_2), \dots, E(x_\mu)] \tag{30}$$

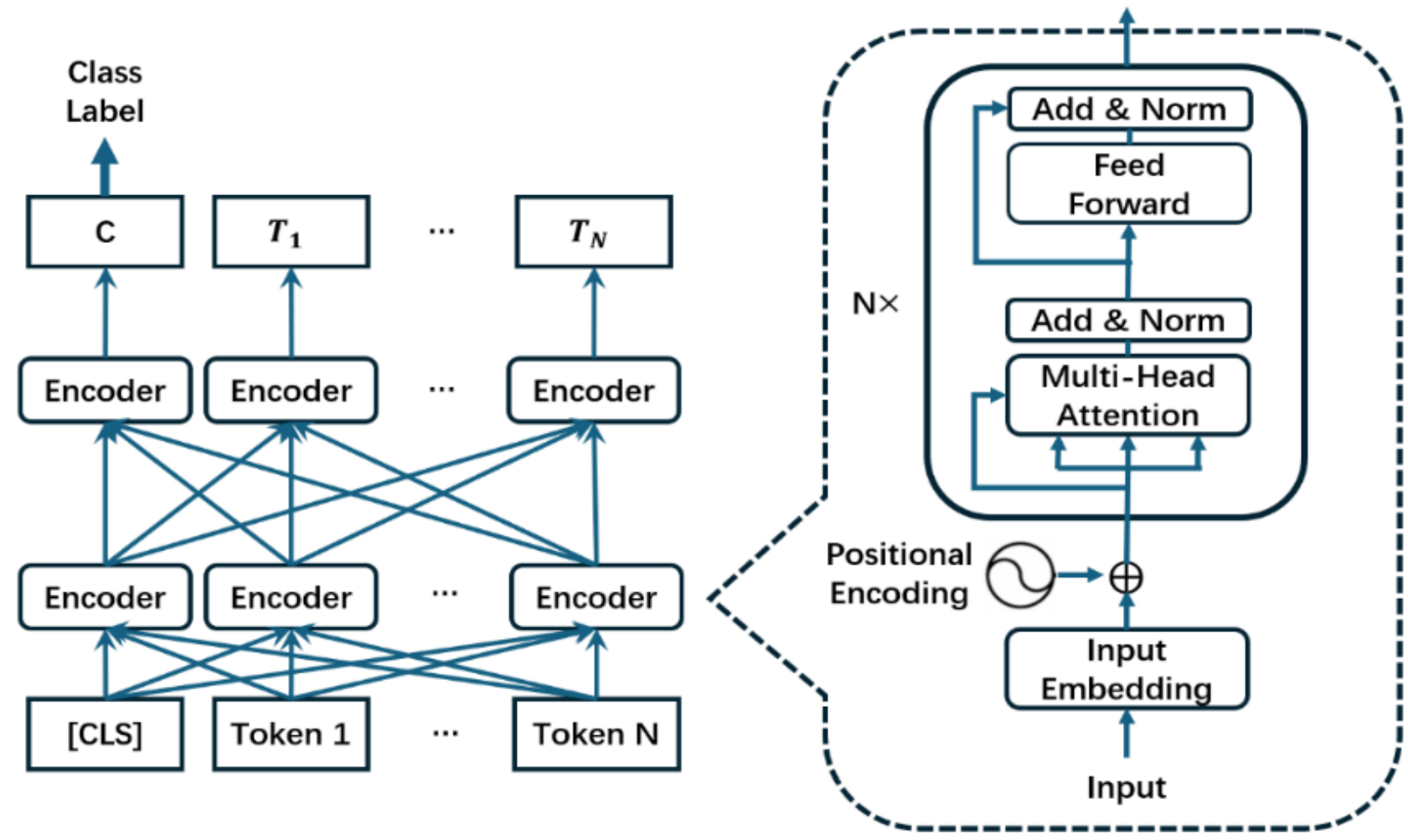

**Figure 4.** Example of ALBERT

Each layer of ALBERT's encoder contains two main operations: Multi-Head Self-Attention and a Feed-Forward Network. In Multi-Head Self-Attention, each token undergoes a linear transformation through $Q$, $K$, and $V$ (Query, Key, and Value). Let $W_Q$ be the weight of $Q$, $W_K$ be the weight of $K$, and $W_V$ be the weight of $V$. The $Q$, $K$, and $V$ are calculated as follows:

$$Q = E(x)W_Q \tag{31}$$

$$K = E(x)W_K \tag{32}$$

$$V = E(x)W_V \tag{33}$$

$d_k$ is the dimension of the head. The attention weight of each head is calculated as follows:

$$Attention(Q, K, V) = \text{softmax}\left(\frac{QK^T}{\sqrt{d_k}}\right)V \tag{34}$$

Let $h$ be the number of heads, $head_i$ be the $i$-th attention head, and $W_m$ be the weights of the multi-head attention. $Concat$ is used to combine multiple attention heads.

$$MultiHead(Q, K, V) = Concat(head_1, head_2, \ldots, head_h)W_m \tag{35}$$

Residual connections and layer normalization are added at each step. Thus, the output $O'$ is:

$$O' = LayerNorm(E(x) + MultiHead(Q, K, V)) \tag{36}$$

The FFN (Feedforward Neural Network) consists of two linear layers plus an activation function (e.g., ReLU), and the output, $O''$, of the Feedforward Neural Network layer is calculated as follows:

$$O'' = LayerNorm\big(O' + FFN(O')\big) \tag{37}$$

Finally, the hidden state of the first token $[CLS]$, $H_{CLS}$, is the representation of the entire sentence, and a fully connected layer on the $[CLS]$ vector is added to form a classification layer. $W_{class}$ is the weight of the classification layer, and $b_{class}$ is the bias of the classification layer. The classification logits are calculated as follows:

$$logits = W_{class} \cdot H_{CLS} + b_{class} \tag{38}$$

## 5. CPC for Evaluation Criteria and Weighted Selection Matrix

### 5.1. Evaluation Criteria

The evaluation criteria for sentiment analysis include Accuracy, F1 score, Precision, and Recall, Specificity [31], Matthews Correlation Coefficient [32], and Cohen's Kappa [33]. Table 6 shows the detailed equations for the evaluation criteria. The notations are defined below.

- $TP$: The number of positive samples successfully identified as positive.
- $FP$: The number of negative samples wrongly classified as positive.
- $FN$: The number of positive samples wrongly classified as negative.
- $TN$: The number of negative samples successfully identified as negative.
- $p_o$: The proportion of samples where the model's prediction matches the true label.

- $p_e$: The expected ratio where the predicted label matches the true label.

**Table 6.** Sentiment Analysis Evaluation Criteria

| Criteria | Definition | Equations |
|---|---|---|
| Accuracy | Accuracy is the proportion of correctly predicted samples to the total sample size. | $\frac{TP+TN}{TP+TN+FP+FN}$ |
| Precision | Precision is the proportion of samples predicted as positive that are actually positive. | $\frac{TP}{TP+FP}$ |
| Recall | Recall is the proportion of all actual positive samples that are correctly predicted as positive. | $\frac{TP}{TP+FN}$ |
| F1-score | F1-score is the harmonic mean of precision and recall, ranging between 0 and 1. | $\frac{2*Precision*Recall}{Precision+Recall}$ |
| Specificity | Specificity is the opposite of recall. This true negative rate quantifies how effectively the negative class is predicted. | $\frac{TN}{TN+FP}$ |
| Matthews Correlation Coefficient | MCC is an indicator for evaluating classification models, considering all four elements of the confusion matrix, and is applicable to imbalanced data. | $\frac{TP*TN-FP*FN}{\sqrt{(TP+FP)(TP+FN)(TN+FP)(TN+FN)}}$ |
| Cohen's Kappa | Kappa measures the degree of consistency between classification results and random chance, accounting for accidental agreement. | $\frac{p_o-p_e}{1-p_e}$ |

Efficiency measured by computational time is a criterion considered in this study. The running time is the sum of the time for training, validation (without 10-fold cross-validation), and testing. Less time consumed indicates a better model. Therefore, efficiency is calculated using reverse Min-Max normalization. Suppose the set of running times for all classification baseline models is $T$, where $\max(T)$ is the longest running time and $\min(T)$ is the shortest running time among them. The running time is normalized to the range of 0 to 1. The running time of the $i$-th model, $T_i$, is normalized as follows:

$$Efficiency = \frac{\max(T)-T_i}{\max(T)-\min(T)} \tag{39}$$

### 5.2. Cognitive pairwise comparison

Cognitive pairwise comparison [9] [10] [11], as the interface for the interaction between humans and machine learning algorithms, is the core technology for determining feature levels. The weight of each feature is presented through the Pairwise Opposite Matrix (POM). Suppose the ideal utility set is $V=\{v_1,\dots,v_n\}$. $b_{ij}=v_i-v_j$ represents the comparison score of the difference between the priorities of two features. $\tilde{B}=[v_i-v_j]$ is an ideal pairwise matrix. The subjective judgment of the pairwise opposite matrix using pairwise interval scales is $B=[b_{ij}]$. $\tilde{B}$ is determined by $B$ as follows:

$$\tilde{B} = [\tilde{b}_{ij}] = [v_i - v_j] \cong [b_{ij}] = B \tag{40}$$

The value $b_{ij}$ is selected from the paired assessment scale $\left\{-\frac{8}{\kappa}, \dots, -\frac{1}{\kappa}, 0, \frac{1}{\kappa}, \dots, \frac{8}{\kappa}\right\}$, indicating {"Extremely less important than", …, "Weakly less important than", "Equal to", "Weakly more important than", …, "Extremely more important than"}. The normal utility $\kappa$ represents the average value of the feature weight (by default, the value of $\kappa$ is 8).

**Table 7.** POM, weight, and rank for evaluation criteria (without Efficiency)

| POM | Accuracy | Precision | Recall | F1 | Specificity | MCC | Kappa | Weight | Rank |
|---|---|---|---|---|---|---|---|---|---|
| Accuracy | 0 | -2 | -3 | -5 | -1 | -8 | -6 | 0.079 | 7 |
| Precision | 2 | 0 | -1 | -3 | 2 | -7 | -4 | 0.115 | 5 |
| Recall | 3 | 1 | 0 | -2 | 3 | -6 | -4 | 0.130 | 4 |
| F1-score | 5 | 3 | 2 | 0 | 4 | -5 | -2 | 0.161 | 3 |
| Specificity | 1 | -2 | -3 | -4 | 0 | -8 | -6 | 0.087 | 6 |
| MCC | 8 | 7 | 6 | 5 | 8 | 0 | 3 | 0.237 | 1 |
| Kappa | 6 | 4 | 4 | 2 | 6 | -3 | 0 | 0.191 | 2 |

**Table 8.** POM, weight, and rank for evaluation criteria (with Efficiency)

| POM | Accuracy | Precision | Recall | F1 | Specificity | MCC | Kappa | Efficiency | Weight | Rank |
|---|---|---|---|---|---|---|---|---|---|---|
| Accuracy | 0 | -2 | -3 | -4 | -1 | -7 | -5 | 4 | 0.090 | 7 |
| Precision | 2 | 0 | -1 | -2 | 1 | -5 | -3 | 4 | 0.117 | 5 |
| Recall | 3 | 1 | 0 | -1 | 2 | -4 | -2 | 5 | 0.133 | 4 |
| F1 | 4 | 2 | 1 | 0 | 3 | -3 | -1 | 7 | 0.150 | 3 |
| Specificity | 1 | -1 | -2 | -3 | 0 | -6 | -4 | 4 | 0.104 | 6 |
| MCC | 7 | 5 | 4 | 3 | 6 | 0 | 2 | 8 | 0.193 | 1 |
| Kappa | 5 | 3 | 2 | 1 | 4 | -2 | 0 | 8 | 0.166 | 2 |
| Efficiency | -4 | -4 | -5 | -7 | -4 | -8 | -8 | 0 | 0.047 | 8 |

Chicco et al. [34] and Sokolova & Lapalme [35] conducted a comparative analysis of different evaluation criteria and concluded that MCC and Kappa are the most recommended evaluation criteria in multi-classification problems, but Kappa's interpretability is inferior. Precision, Recall, and F1-score are complementary criteria, but they only focus on the positive class and should not be used alone. Specificity, the counterpart of Recall, is used to measure the ability to identify negative classes, but it should be used together with Recall. Accuracy is highly likely to be misleading in imbalanced data and should not be the main evaluation criterion. The influence of the time factor is smaller than that of the above seven evaluation criteria. Based on the conclusions, two POMs for comparative evaluation criteria are presented in Table 7 and Table 8, respectively.

It is important to clarify that the proposed CPC-CMS framework is open and adaptable to any decision-maker's preference profile. In this study, to establish a standardized baseline evaluation, the primary Pairwise

Opposite Matrices (POMs) in Table 7 and Table 8 are constructed as a Literature-Grounded Reference Scenario. Rather than reflecting ad-hoc author choices, the relative importance between criteria is systematically derived from theoretical and empirical insights established in multi-classification benchmarking literature [34] [35].

The POM example (without Efficiency) for the simulation in Section 6 is shown in Table 7. For example, the score -2 for comparing Accuracy and Precision means that Accuracy is -2 units more important than Precision (i.e., Precision is 2 units more important than Accuracy). The mathematical form is $v_{Accuracy} - v_{Precision} = -2$. The POM example (with Efficiency) for the simulation in Section 6 can also be generated in the same way, and the details are presented in Table 8. The cognitive pairwise matrix $B$ is verified by the Accordance Index ($AI$) as shown as follows.

$$AI = \frac{1}{n^2}\sum_{i=1}^{n}\sum_{j=1}^{n}\sqrt{\frac{1}{n}\sum_{p=1}^{n}\left(\frac{\left(b_{ip}+b_{pj}-b_{ij}\right)}{\kappa}\right)^2} \quad (41)$$

If $AI = 0$, then $B$ is completely consistent; if $AI > 0.1$, $B$ may need to be modified; if $0 < AI \leq 0.1$, $B$ is acceptable. The accordance index of the POM in Table 7 is 0.0747 and that in Table 8 is 0.0647, which are within the acceptable range.

To obtain weights, the Row Average plus the following normal Utility (RAU) is used as the weight operator.

$$v_i = \kappa + \frac{1}{n}\sum_{j=1}^{n} b_{ij}, \forall i \in \{1, \dots, n\} \quad (42)$$

A single weight $v_i$ from the POM is rescaled to a normalized weight vector by the rescaling function (or normalization). The function is as follows:

$$w_i = \frac{v_i}{n\kappa}, \forall i \in \{1, \dots, n\} \quad (43)$$

The weights and ranks for all features are presented in Table 7. To illustrate the calculation, the weight of Accuracy is calculated using Eq. 42 as follows:

$$v_1 = \frac{(0 + (-2) + (-3) + (-5) + (-1) + (-8) + (-6))}{7} + 8 \approx 4.428$$

Thus, the relative weight of Accuracy is calculated using Eq. 43 as follows:

$$w_1 = \frac{4.428}{7 \times 8} \approx 0.079$$

The weights of the rest of the features are calculated using the same method as demonstrated above.

### 5.3. Weighted Decision Matrix

The Weighted Decision Matrix is a selection mechanism based on scoring to select the best among multiple classification candidate models. Suppose there are $m$ classification candidate models and $n$ evaluation criteria. The weight of each evaluation criterion calculated by CPC can form the weight matrix $W$:

$$W = [w_1\ w_2\ \cdots\ w_n] \tag{44}$$

$S$ is the score matrix of the $m$ models on the $n$ evaluation criteria as follows:

$$S = \begin{bmatrix} s_{11} & s_{12} & \cdots & s_{12} \\ s_{21} & s_{22} & \cdots & s_{2n} \\ \vdots & \vdots & \ddots & \vdots \\ s_{m1} & s_{m2} & s_{m3} & s_{mn} \end{bmatrix} \tag{45}$$

Let $S_{ij}$ be the score of the $i$-th model on the $j$-th evaluation criterion, and $w_j$ be the weight of the $j$-th evaluation criterion. Therefore, the comprehensive score, $G_i$, which is the weighted sum of the $i$-th model on all evaluation criteria can be expressed as:

$$G_i = \sum_{j=1}^{n} w_j \cdot S_{ij} \tag{46}$$

The weights of all evaluation criteria (without Efficiency) are in Table 7, and the test results of all classification models on all evaluation criteria (with Efficiency) on the Case 1 dataset are presented in Table 15. To illustrate the calculation, the comprehensive score of the LSVC is calculated using Eq. 46 as follows.

$$G_1 = [0.079\ \ 0.115\ \ 0.130\ \ 0.161\ \ 0.087\ \ 0.237\ \ 0.191] \begin{bmatrix} 0.781 \\ 0.725 \\ 0.762 \\ 0.738 \\ 0.960 \\ 0.718 \\ 0.717 \end{bmatrix} \approx 0.754$$

The comprehensive scores of the remaining classification models are calculated using the same method as demonstrated above. Finally, the one with the highest score is the best model.

## 6. Simulation

### 6.1. Background

For the computational environment of the experiments, the CPU is Intel(R) Xeon(R) CPU @ 2.20GHz, the GPU is an NVIDIA Tesla P100 PCIe 16GB, and Python 3.11.11 is used. The machine learning models are implemented using scikit-learn [36], TensorFlow [37] and Transformers [38].

To demonstrate the usability of the proposed method, three datasets [19] [39] [40] are evaluated. To guarantee a fair comparison, it is imperative that identical data splits are used across all evaluated models. This was enforced by controlling the splitting algorithm with three fixed random seeds (101, 202, and 303). For each seed, the dataset is divided into a training set (80%), a validation set (10%), and a test set (10%) using the *train_test_split* function [36]. Crucially, stratification was applied during this split to preserve the original class distributions across all subsets.

Furthermore, 10-fold cross-validation is performed on the training data to generate the respective models and extract evaluation scores (Accuracy, Precision, Recall, F1, Specificity, MCC, Kappa, and Time Efficiency). Reporting metrics on both the training/validation sets and the unseen test set allows us to actively monitor for overfitting; algorithms exhibiting significant performance drops on the test set are flagged and penalized in the selection system. The final evaluation scores reported are the overall means derived from the 10-fold cross-validation repeated across the three stratified seed configurations.

**Table 9.** A comparison of three Tokenization tools

| Tool | Method | Vocabulary list | Output |
|---|---|---|---|
| word_tokenize [41] | Rules and regular expressions | No | List of words |
| Tokenizer [37] | Based on Spaces and punctuation | Yes (Self-built) | List of word indices |
| AutoTokenizer [38] | Subword tokenization | Yes (Pretrained) | Token IDs ＋ Attention Mask |

**Table 10.** A comparison of three Feature extraction tools

| Tool | Tokenization | Embedding Type | Output |
|---|---|---|---|
| TfidfVectorizer [36] | Word-level | TF-IDF Weighted | Sparse Matrix (Bag-of-Words) |
| Embedding Layer [37] | Word-level | Learned Word Embeddings | Dense Matrix<br>(seq_lenem × bedding_dim) |
| AutoTokenizer [38] | Subword-level | ALBERT (Pretrained) | Token IDs + Attention Mask |

As detailed in Section 3, the document dataset requires preprocessing and feature extraction tailored to each model architecture. Initial text cleaning is performed using Python's built-in *re* module following the steps outlined in Section 3.2. As summarized in Table 9 and Table 10, distinct libraries, including NLTK [41], scikit-learn [36], TensorFlow [37], and Transformers [38], are employed to generate the specific input formats required by each model paradigm. Specifically, the LSTM network is implemented using *Tokenizer* [37] and an *Embedding Layer* [37], whereas ALBERT utilizes a pretrained subword-level *AutoTokenizer* [38]. Traditional machine learning algorithms (Naive Bayes, Random Forest, LSVC, Logistic Regression, and XGBoost) utilize *word_tokenize* [41] and *PorterStemmer* [41] for preprocessing, alongside *TfidfVectorizer* [36] for feature extraction. Evaluating models as complete end-to-end pipelines ensures that the multi-criteria decision framework captures the true operational trade-offs, such as feature extraction execution time and predictive performance, inherent to each complete technical stack.

### 6.2. Case 1

For Case 1, the dataset [19] is consolidated and cleaned from different platforms, including social media posts, Twitter posts, and Reddit posts, for performing sentiment analysis. This dataset contains 52,681 documents and divides mental health into seven states: Normal (16,343 documents), Depression (15,404 documents), Suicidal (10,652 documents), Anxiety (3,841 documents), Bipolar (2,777 documents), Stress (2,587 documents), and Personality disorder (1,077 documents). Personality disorder, Stress, Bipolar, and Anxiety account for 2.0%, 4.9%, 5.3%, and 7.3% of the total instances of the dataset, respectively, while Normal, Depression, and Suicidal account for 31.0%, 29.2%, and 20.2% of the total instances, respectively. The dataset is imbalanced. A class weighting sampling technique [36] is used to balance the dataset for better training effects.

In Table 11, the seven mental states contained in the Case 1 dataset are presented, and each document corresponds to one mental state. As each raw document contains many repetitive words and irregular punctuation marks, the documents require pre-processing and transformation into numerical features that can be used to train the classification models, using the tools mentioned in Section 6.1. Seven baseline classification models are used to determine the mental state (Normal, Depression, Suicidal, Anxiety, Bipolar, Stress, or Personality disorder) corresponding to each document. Their classification performance and time consumption are compared based on eight weighted evaluation criteria to select the best model.

**Table 11.** Example of documents in Case 1 dataset

| **Document** | **Sentiment** |
|---|---|
| It's worth not buying an airpod pro. But already bought. Huwaaa | Normal |
| Just feeling real low today or atleast for the past few hours. It sucks. Hope you are alright Okay | Depression |
| Looking for fast and effective way to end all of these. Ideas? Fast and effective way | Suicidal |
| My life is chaos. There is no solution. Fear of the uncertain. Restless direction. | Anxiety |
| Current status. Manic. Sleep deprived. Tired but i cant sleep. | Bipolar |
| stress I lost my wallet... so stress. Ok goodnight. And i hope to find it tmr.. | Stress |
| Friends? Does anyone look for a friend? Wanna be friends? | Personality disorder |

**Table 12.** The size of data subsets after the train-validation-test split on Case 1 dataset

| **Datasets** | **Normal** | **Depression** | **Suicidal** | **Anxiety** | **Bipolar** | **Stress** | **Personality disorder** | **Total** |
|---|---|---|---|---|---|---|---|---|
| Training | 13106<br>(80.2%) | 12354<br>(80.2%) | 8468<br>(79.5%) | 3073<br>(80.0%) | 2211<br>(79.6%) | 2057<br>(79.5%) | 875<br>(81.2%) | 42144<br>(80%) |
| Validation | 1618<br>(9.9%) | 1525<br>(9.9%) | 1092<br>(10.3%) | 384<br>(10.0%) | 283<br>(10.2%) | 265<br>(10.2%) | 101<br>(9.4%) | 5268<br>(10%) |
| Test | 1619<br>(9.9%) | 1525<br>(9.9%) | 1092<br>(10.3%) | 384<br>(10.0%) | 283<br>(10.2%) | 265<br>(10.2%) | 101<br>(9.4%) | 5269<br>(10%) |
| Total | 16343<br>(31.0%) | 15404<br>(29.2%) | 10652<br>(20.2%) | 3841<br>(7.3%) | 2777<br>(5.3%) | 2587<br>(4.9%) | 1077<br>(2.0%) | 52681<br>(100%) |

The size of data subsets after the train-validation-test split is shown in Table 12. A total of 42,144 documents (80% of the total) are used for training classification models, 5,268 documents (10% of the total) are used to verify the models, and 5,269 documents (10% of the total) are used to test the performance of the models. Parameter settings for the baseline models are shown in Table 13.

**Table 13.** Parameter settings for the baseline models for Case 1

| Model | Methods and Parameters |
|---|---|
| **LSVC** | LinearSVC(C=1.0, penalty='l1', loss='squared_hinge', dual=False, random_state=101) |
| **Bernoulli Naïve Bayes** | BernoulliNB(alpha=0.1, binarize=0.0) |
| **Random Forest** | RandomForestClassifier(n_estimators=100, random_state=42) |
| **Logistic Regression** | LogisticRegression(solver='liblinear', penalty='l1', C=10, random_state=101) |
| **XGBoost** | XGBClassifier(learning_rate=0.2, max_depth=7, n_estimators=500, random_state=101,<br>tree_method='gpu_hist') |
| **LSTM** | MAX_LEN = 128<br>VOCAB_SIZE = 50000<br>model = Sequential()<br>model.add(Embedding(input_dim=VOCAB_SIZE, output_dim=64,input_length=MAX_LEN))<br>model.add(LSTM(64, return_sequences=False, kernel_regularizer=regularizers.l2(0.001)))<br>model.add(Dropout(0.5))<br>model.add(Dense(NUM_CLASSES, activation='softmax'))<br>model.compile(loss='categorical_crossentropy', optimizer = Adam(learning_rate=0.001)<br>metrics=['accuracy'])<br>early_stop = EarlyStopping(monitor='val_loss', patience=5, restore_best_weights=True)<br>model_checkpoint = ModelCheckpoint("best_lstm_model.h5", monitor='val_accuracy',<br>save_best_only=True)<br>def lr_scheduler(epoch, lr): if epoch < 15: return float(lr) else: return float(lr * np.exp(-0.1))<br>lr_callback = LearningRateScheduler(lr_scheduler, verbose=1)<br>history = model.fit(X_train_pad, y_train_cat, validation_data=(X_val_pad, y_val_cat), epochs=30,<br>batch_size=32, callbacks=[early_stop, model_checkpoint, lr_callback], verbose=1,<br>class_weight=class_weights_dict) |
| **ALBERT** | training_args = TrainingArguments(output_dir="./transformer_results", num_train_epochs=30,<br>per_device_train_batch_size=16, per_device_eval_batch_size=16, eval_strategy="epoch",<br>save_strategy="epoch", logging_dir='./logs', logging_strategy="epoch",<br>logging_steps=1,save_total_limit=2, learning_rate=3e-5, warmup_steps=1000, weight_decay=0.01,<br>load_best_model_at_end=True, metric_for_best_model='accuracy',<br>report_to='none', fp16=torch.cuda.is_available(), gradient_accumulation_steps=4, seed=42)<br>early_stopping_callback = EarlyStoppingCallback(early_stopping_patience=3)<br>trainer = Trainer(model=model, args=training_args, train_dataset=tokenized_train,<br>eval_dataset=tokenized_val, tokenizer=tokenizer, compute_metrics=compute_metrics,<br>callbacks=[metrics_callback, early_stopping_callback]) |

As LSTM and ALBERT have high complexity, a large number of parameters, and a high dependence on training data, they are prone to overfit on small and medium-sized data. Figure 5 shows the overfitting of LSTM with 30 training epochs, even though dropout, L2 regularization, learning rate scheduling, and weight decay are used to attempt to prevent overfitting. To prevent overfitting, the early stopping mechanism is used, and the training step stops at Epoch 14, as shown in Figure 6. The early stopping mechanism not only saves training time but also preserves the weights of the best-performing models on the validation set for testing their performance on the test set.

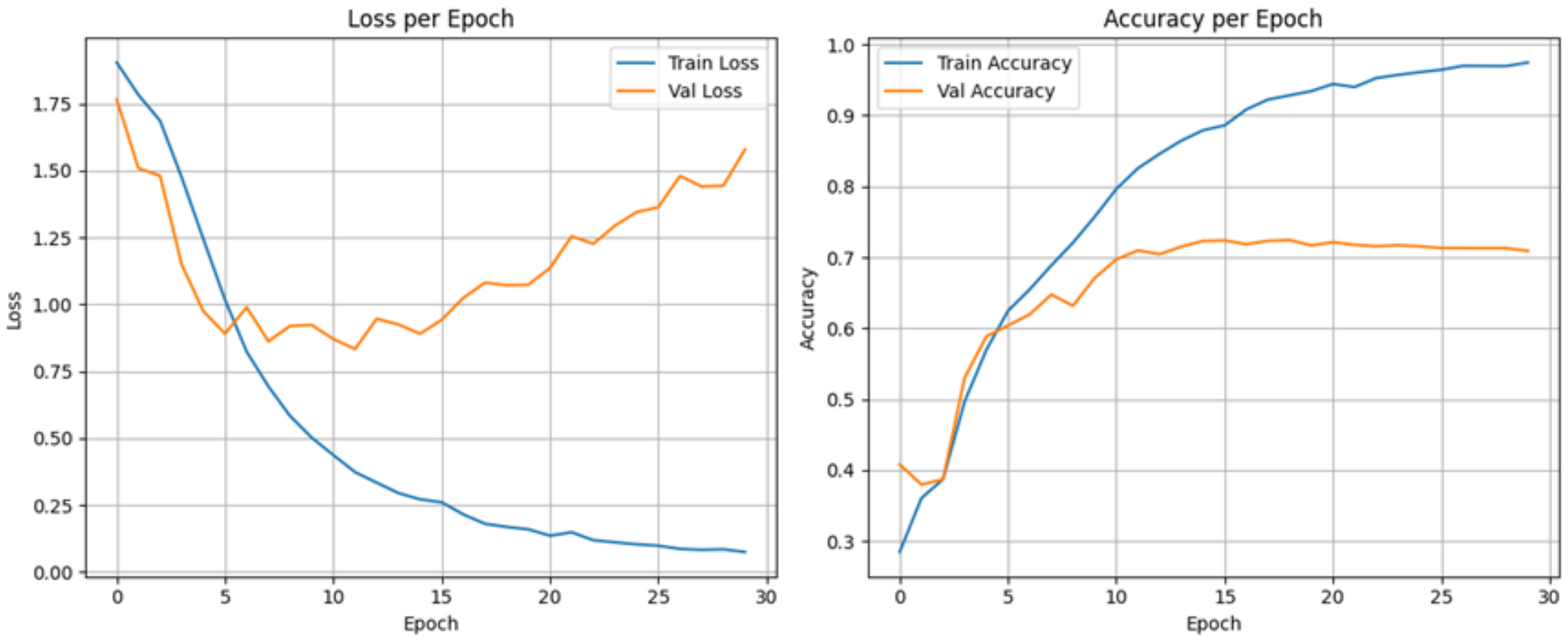

**Figure 5.** Overfitting of LSTM

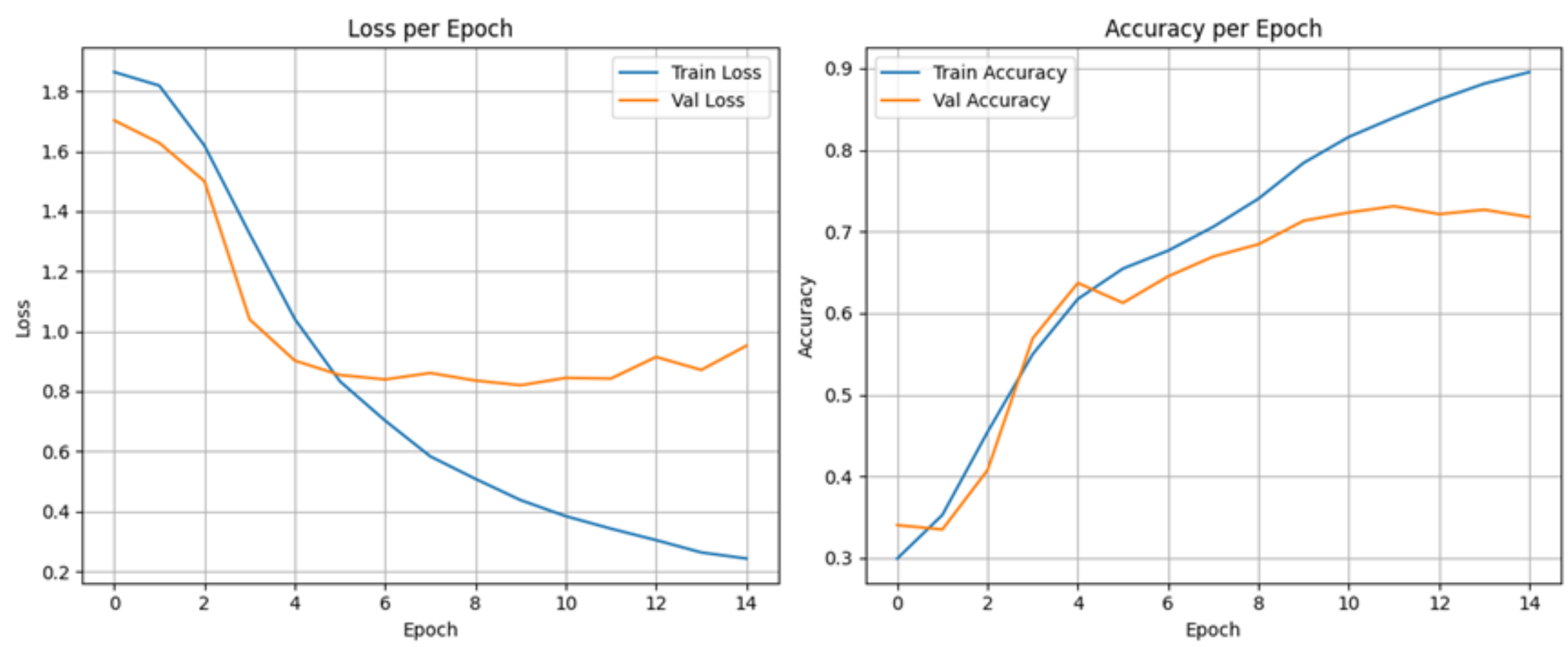

**Figure 6.** Early stopping for LSTM

**Table 14.** The evaluation results of 10-fold cross-validation on Case 1 validation dataset

| **Model** | **Accuracy** | **Precision** | **Recall** | **F1-score** | **Specificity** | **MCC** | **Kappa** |
|---|---|---|---|---|---|---|---|
| LSVC | 0.781±0.001 | 0.736±0.003 | 0.764±0.001 | 0.748±0.002 | 0.960±0.000 | 0.717±0.001 | 0.716±0.001 |
| Bernoulli Naïve Bayes | 0.647±0.002 | 0.639±0.004 | 0.602±0.004 | 0.607±0.003 | 0.934±0.000 | 0.546±0.003 | 0.537±0.003 |
| Random Forest | 0.695±0.002 | 0.838±0.001 | 0.510±0.009 | 0.572±0.011 | 0.939±0.000 | 0.600±0.003 | 0.583±0.003 |
| Logistic Regression | 0.769±0.004 | 0.765±0.002 | 0.725±0.002 | 0.742±0.002 | 0.957±0.001 | 0.697±0.005 | 0.697±0.006 |
| XGBoost | 0.808±0.006 | 0.825±0.008 | 0.761±0.010 | 0.788±0.010 | 0.964±0.001 | 0.748±0.008 | 0.748±0.009 |
| LSTM | 0.728±0.008 | 0.740±0.006 | 0.826±0.006 | 0.723±0.008 | 0.950±0.004 | 0.648±0.008 | 0.647±0.009 |
| ALBERT | 0.834±0.002 | 0.836±0.001 | 0.834±0.002 | 0.835±0.002 | 0.970±0.001 | 0.781±0.002 | 0.781±0.002 |

Through the three run comparisons described in Section 6.1, evaluation results are obtained based on 10-fold cross-validation, as shown in Table 14. Since the fluctuations (error range) under multiple training runs of the same partitioning strategy are generally within ±0.01, all models appear to have stable performance. The performance of the models can be further tested on the test set.

**Table 15.** The evaluation results on Case 1 testing dataset

| Model | Accuracy | Precision | Recall | F1-score | Specificity | MCC | Kappa |
|---|---|---|---|---|---|---|---|
| LSVC | 0.781 | 0.725 | 0.762 | 0.738 | 0.960 | 0.718 | 0.717 |
| Bernoulli Naive Bayes | 0.643 | 0.627 | 0.596 | 0.594 | 0.933 | 0.542 | 0.533 |
| Random Forest | 0.704 | 0.845 | 0.527 | 0.594 | 0.941 | 0.610 | 0.596 |
| Logistic Regression | 0.766 | 0.750 | 0.732 | 0.740 | 0.956 | 0.693 | 0.693 |
| XGBoost | 0.808 | 0.823 | 0.761 | 0.787 | 0.964 | 0.748 | 0.748 |
| LSTM | 0.727 | 0.744 | 0.727 | 0.729 | 0.951 | 0.658 | 0.649 |
| ALBERT | 0.821 | 0.824 | 0.821 | 0.821 | 0.967 | 0.767 | 0.766 |

**Table 16.** The Efficiency result of the model on Case 1 dataset

| | LSVC | Bernoulli Naive Bayes | Random Forest | Logistic Regression | XGBOOST | LSTM | ALBERT |
|---|---|---|---|---|---|---|---|
| Running time | 120.006s | 0.683s | 389.995s | 65.654s | 151.294s | 225.221s | 4052.547s |
| Efficiency | 0.971 | 1.000 | 0.904 | 0.984 | 0.963 | 0.945 | 0.000 |

**Table 17.** The CPC-CMS results on Case 1 dataset

| **Results without considering the efficiency factor** | | | | | | | |
|---|---|---|---|---|---|---|---|
| | **LSVC** | **Bernoulli Naive Bayes** | **Random Forest** | **Logistic Regression** | **XGBoost** | **LSTM** | **ALBERT** |
| **Score** | 0.754 | 0.607 | 0.657 | 0.741 | 0.788 | 0.718 | 0.811 |
| **Rank** | 3 | 7 | 6 | 4 | 2 | 5 | 1 |
| **Results with considering the efficiency factor** | | | | | | | |
| | **LSVC** | **Bernoulli Naive Bayes** | **Random Forest** | **Logistic Regression** | **XGBoost** | **LSTM** | **ALBERT** |
| **Score** | 0.770 | 0.637 | 0.678 | 0.760 | 0.802 | 0.737 | 0.778 |
| **Rank** | 3 | 7 | 6 | 4 | 1 | 5 | 2 |

The evaluation results on the test set are shown in Table 15. For precision, Logistic Regression performs better than LSVC; for accuracy, LSVC performs better than Logistic Regression. For the seven evaluation criteria, ALBERT is the best, but ALBERT takes the longest time and has the lowest efficiency, as shown in Table 16. Therefore, the importance of all eight evaluation criteria needs to be comprehensively considered. The final score and the ranking are calculated by using the weights of each evaluation criterion in Table 7 and Table 8, respectively.

Table 17 shows the results of CPC-CMS. Considering the weighted aggregate results of all evaluation criteria, LSVC performs better than Logistic Regression. If the time factor is not considered, ALBERT is the best model. However, if the time factor is considered, the rankings of the other models remain unchanged, while XGBoost rises from second to first place, outperforming ALBERT.

### 6.3. Case 2

For Case 2, the dataset [39] is a sentiment dataset from Twitter. While the "Irrelevant" category, which does not contain sentiment, is excluded, 61,120 documents are classified into three categories: Positive (20,654 documents), Neutral (18,108 documents), and Negative (22,358 documents). As the three categories account for 33.8%, 29.6% and 36.6% of the total instances in the dataset, respectively, the dataset is balanced.

In Table 18, an example from the Case 2 dataset, containing three sentiments, is presented, and each document corresponds to one sentiment. The raw documents are preprocessed and transformed into numerical features. The process of simulation is similar to the process in Case 1, and the goal is to select the best model.

**Table 18**. Example of documents in Case 2 dataset

| Document | Sentiment |
|---|---|
| I can't wait to play the beta this weekend! Oh wait.... . I have an Xbox. . . Never mind. | Positive |
| My friends,. Prize draw tomorrow live at 12pm CET . . Have a great day . | Neutral |
| The hell is with Nvidia all of a sudden. . . Why is their new line up so cheap. What..? | Negative |

**Table 19.** The size of data subsets after the train-validation-test split on Case 2 dataset

| Datasets | Positive | Neutral | Negative | Total |
|---|---|---|---|---|
| Training | 16524 (80.0%) | 14486 (80.0%) | 17886 (80.0%) | 48896 (80%) |
| Validation | 2065 (10.0%) | 1811 (10.0%) | 2236 (10.0%) | 6112 (10%) |
| Test | 2065 (10.0%) | 1811 (10.0%) | 2236 (10.0%) | 6112 (10%) |
| Total | 20654 (33.8%) | 18108 (29.6%) | 22358 (36.6%) | 61120 (100%) |

**Table 20.** Parameter settings for baseline models in Case 2

| Model | Methods and Parameters |
|---|---|
| **LSTM** | *MAX_LEN = 128*<br>*VOCAB_SIZE = 50000*<br>*model = Sequential()*<br>*model.add(Embedding(input_dim=VOCAB_SIZE, output_dim=128, input_length=MAX_LEN))*<br>*model.add(Bidirectional(LSTM(128, return_sequences=False, kernel_regularizer=regularizers.l2(0.001))))*<br>*model.add(BatchNormalization())*<br>*model.add(Dropout(0.4))*<br>*model.add(Dense(NUM_CLASSES, activation='softmax'))*<br>*model.compile(loss='categorical_crossentropy',optimizer=Adam(learning_rate=0.001), metrics=['accuracy'])*<br>*early_stop =EarlyStopping(monitor='val_loss',patience=5,restore_best_weights=True)*<br>*model_checkpoint = ModelCheckpoint("best_lstm_model.h5",monitor='val_accuracy',*<br>*save_best_only=True)*<br>*def lr_scheduler(epoch, lr):if epoch < 5:return float(lr) else:return float(lr * np.exp(-0.1))*<br>*lr_callback = LearningRateScheduler(lr_scheduler, verbose=1)*<br>*history = model.fit(X_train_pad, y_train_cat, validation_data=(X_val_pad, y_val_cat), epochs=30, batch_size=32,*<br>*callbacks=[early_stop, model_checkpoint, lr_callback], verbose=1, class_weight=class_weights_dict)* |

The size of data subsets after the train-validation-test split is shown in Table 19. The training set contains

48,896 documents (80% of the total), the validation set contains 6,112 documents (10% of the total), and the test set contains 6,112 documents (10% of the total). The processes of training, validation, and testing are similar to those in Case 1. Overfitting also occurred, and thus an early stopping mechanism was added. Parameter settings for the baseline models are the same as those in Case 1, but the LSTM settings in Case 2 are shown in Table 20.

According to the results of 10-fold cross-validation shown in Table **21,** the classification performance of all models on the validation set is stable, with a fluctuation of less than 0.006, making it even more stable than in Case 1. The evaluation results on the test set and time consumption are shown in Table 22 and Table 23, respectively. ALBERT ranks first based on all seven evaluation criteria, but its advantages are not obvious, and its running time is still the longest. In contrast, many traditional machine learning models perform equally well, especially Random Forest. To select the best model, CPC-CMS is used to obtain the final scores.

Table 24 shows the results of CPC-CMS. ALBERT is the best model in terms of performance, and Random Forest follows closely behind, with a difference of only 0.013. After considering the time factor, the ranking changes. Random Forest becomes the best model, and ALBERT ranks third. Random Forest excels in performance and time consumption when dealing with documents like those in the Case 2 dataset.

**Table 21.** The evaluation results of 10-fold cross-validation on Case 2 validation dataset

| Model | Accuracy | Precision | Recall | F1-score | Specificity | MCC | Kappa |
|---|---|---|---|---|---|---|---|
| LSVC | 0.883±0.001 | 0.882±0.000 | 0.882±0.000 | 0.882±0.000 | 0.941±0.001 | 0.824±0.001 | 0.824±0.001 |
| Bernoulli Naïve Bayes | 0.857±0.002 | 0.869±0.002 | 0.856±0.002 | 0.857±0.003 | 0.928±0.001 | 0.790±0.003 | 0.784±0.003 |
| Random Forest | 0.914±0.000 | 0.914±0.000 | 0.912±0.000 | 0.913±0.000 | 0.957±0.000 | 0.870±0.000 | 0.870±0.000 |
| Logistic Regression | 0.889±0.001 | 0.888±0.001 | 0.888±0.001 | 0.888±0.001 | 0.945±0.001 | 0.833±0.002 | 0.833±0.002 |
| XGBoost | 0.834±0.001 | 0.835±0.001 | 0.830±0.001 | 0.831±0.001 | 0.916±0.001 | 0.750±0.001 | 0.749±0.002 |
| LSTM | 0.900±0.004 | 0.901±0.003 | 0.900±0.004 | 0.900±0.004 | 0.950±0.002 | 0.850±0.006 | 0.849±0.006 |
| ALBERT | 0.931±0.001 | 0.930±0.002 | 0.931±0.001 | 0.931±0.001 | 0.966±0.001 | 0.897±0.001 | 0.897±0.002 |

**Table 22.** The evaluation results on Case 2 testing dataset

| Model | Accuracy | Precision | Recall | F1-score | Specificity | MCC | Kappa |
|---|---|---|---|---|---|---|---|
| LSVC | 0.883 | 0.882 | 0.882 | 0.882 | 0.941 | 0.823 | 0.823 |
| Bernoulli Naive Bayes | 0.819 | 0.834 | 0.817 | 0.818 | 0.909 | 0.735 | 0.727 |
| Random Forest | 0.917 | 0.917 | 0.916 | 0.917 | 0.958 | 0.876 | 0.876 |
| Logistic Regression | 0.886 | 0.884 | 0.884 | 0.884 | 0.943 | 0.828 | 0.828 |
| XGBoost | 0.837 | 0.839 | 0.833 | 0.834 | 0.918 | 0.755 | 0.754 |
| LSTM | 0.897 | 0.897 | 0.897 | 0.897 | 0.948 | 0.845 | 0.844 |
| ALBERT | 0.929 | 0.929 | 0.929 | 0.929 | 0.964 | 0.893 | 0.893 |

**Table 23.** The Efficiency result of the model on Case 2 dataset

| | LSVC | Bernoulli Naive Bayes | Random Forest | Logistic Regression | XGBoost | LSTM | ALBERT |
|---|---|---|---|---|---|---|---|
| Running time | 16.832s | 0.324s | 251.759s | 5.181s | 31.860s | 334.561s | 12041.739s |
| Efficiency | 0.999 | 1.000 | 0.979 | 1.000 | 0.997 | 0.972 | 0.000 |

**Table 24.** CPC-CMS results on Case 2 dataset

| **Results without considering the efficiency factor** | | | | | | | |
|---|---|---|---|---|---|---|---|
| | **LSVC** | **Bernoulli Naive Bayes** | **Random Forest** | **Logistic Regression** | **XGBoost** | **LSTM** | **ALBERT** |
| **Score** | 0.862 | 0.791 | 0.903 | 0.865 | 0.808 | 0.879 | 0.917 |
| **Rank** | 5 | 7 | 2 | 4 | 6 | 3 | 1 |
| **Results with considering the efficiency factor** | | | | | | | |
| | **LSVC** | **Bernoulli Naive Bayes** | **Random Forest** | **Logistic Regression** | **XGBoost** | **LSTM** | **ALBERT** |
| **Score** | 0.873 | 0.807 | 0.909 | 0.876 | 0.823 | 0.887 | 0.876 |
| **Rank** | 5 | 7 | 1 | 4 | 6 | 2 | 3 |

### 6.4. Case 3

For Case 3, the dataset [40] contains data from two sources, Twitter and Reddit, but only the data from Reddit is considered in this study. A total of 37,149 documents in the Reddit dataset are categorized into three sentiment types: Negative (8,277 documents), Neutral (13,042 documents), and Positive (15,830 documents), accounting for 22.3%, 35.1%, and 42.6%, respectively. Data balancing (using a class weighting sampling technique [36]) is required.

Three sample documents are presented in Table 25, and each document corresponds to one sentiment. The sentiments in the Case 3 dataset have been digitized (1 for positive, 0 for neutral, and -1 for negative). The documents need to be further preprocessed and transformed. Since the sentiment labels are the same as those in Case 2, the simulation process and the purpose of selecting the best model are the same.

**Table 25.** Example of documents in Case 3 dataset

| **Document** | **Sentiment** |
|---|---|
| dont worry about trying explain yourself just meditate regularly and try hard you can more aware everything else will follow coming from someone who has been throught his situation welcome pms | 1 |
| how devarshi patel the ultimate rockstar from ahmadabad | 0 |
| how the public transportation gujarat from friends experience really bad when compared tamil nadu andhra pradesh | -1 |

The size of data subsets after the train-validation-test split is shown in Table 26. There are 29,719 documents (80% of the total) in the training set (80% of the total), 3,715 documents (10% of the total) in the validation set, and 3,715 documents (10% of the total) in the test set. The processes of training, validation, and testing are the same as those in Case 1. Overfitting also occurred, and the same handling methods as in Case 1 were used. Table 27 presents the differences from Case 1 in terms of models and parameters.

**Table 26.** The size of data subsets after the train-validation-test split on Case 3 dataset

| Datasets | Positive | Neutral | Negative | Total |
|---|---|---|---|---|
| Training | 12664 (80.0%) | 10432 (80.0%) | 6623 (80.0%) | 29719 (80%) |
| Validation | 1583 (10.0%) | 1305 (10.0%) | 827 (10.0%) | 3715 (10%) |
| Test | 1583 (10.0%) | 1305 (10.0%) | 827 (10.0%) | 3715 (10%) |
| Total | 15830 (42.6%) | 13042 (35.1%) | 8277 (22.3%) | 37149 (100%) |

The evaluation results shown in Table 28 are obtained based on 10-fold cross-validation. All models still demonstrate stable performance on the validation set. The evaluation results of the test set and time consumption are shown in Table 29 and

Table **30**, respectively. ALBERT has the best performance but the highest time consumption.

**Table 27.** Parameter settings for the baseline models for Case 3

| Model | Methods and Parameters |
|---|---|
| **LSTM** | MAX_LEN = 128<br>VOCAB_SIZE = 50000<br>model = Sequential()<br>model.add(Embedding(input_dim=VOCAB_SIZE, output_dim=64,input_length=MAX_LEN))<br>model.add(LSTM(64,return_sequences=False,kernel_regularizer=regularizers.l2(0.001)))<br>model.add(Dropout(0.5))<br>model.add(Dense(NUM_CLASSES, activation='softmax'))<br>model.compile(loss='categorical_crossentropy', optimizer = Adam(learning_rate=0.001) metrics=['accuracy'])<br>early_stop = EarlyStopping(monitor='val_loss', patience=3, restore_best_weights=True)<br>model_checkpoint = ModelCheckpoint("best_lstm_model.h5", monitor='val_accuracy', save_best_only=True)<br>def lr_scheduler(epoch, lr): if epoch < 5: return float(lr) else:return float(lr * np.exp(0.1))<br>lr_callback = LearningRateScheduler(lr_scheduler, verbose=1)<br>history = model.fit(X_train_pad, y_train_cat, validation_data=(X_val_pad, y_val_cat), epochs=30, batch_size=32, callbacks=[early_stop, model_checkpoint, lr_callback], verbose=1, class_weight=class_weights_dict) |

**Table 28.** The evaluation results of 10-fold cross-validation on Case 3 validation dataset

| Model | Accuracy | Precision | Recall | F1-score | Specificity | MCC | Kappa |
|---|---|---|---|---|---|---|---|
| **LSVC** | 0.898±0.002 | 0.890±0.002 | 0.889±0.002 | 0.889±0.002 | 0.948±0.001 | 0.843±0.002 | 0.842±0.003 |
| **Bernoulli Naïve Bayes** | 0.612±0.001 | 0.606±0.002 | 0.586±0.001 | 0.572±0.001 | 0.801±0.001 | 0.418±0.002 | 0.398±0.002 |
| **Random Forest** | 0.747±0.002 | 0.799±0.001 | 0.674±0.002 | 0.667±0.002 | 0.859±0.001 | 0.613±0.002 | 0.588±0.002 |
| **Logistic Regression** | 0.893±0.001 | 0.887±0.002 | 0.880±0.002 | 0.883±0.002 | 0.945±0.001 | 0.834±0.002 | 0.834±0.003 |
| **XGBoost** | 0.879±0.001 | 0.874±0.001 | 0.860±0.001 | 0.865±0.001 | 0.937±0.000 | 0.813±0.002 | 0.812±0.002 |
| **LSTM** | 0.905±0.002 | 0.905±0.008 | 0.905±0.002 | 0.905±0.004 | 0.952±0.004 | 0.852±0.006 | 0.852±0.006 |
| **ALBERT** | 0.953±0.004 | 0.953±0.003 | 0.953±0.004 | 0.953±0.004 | 0.975±0.001 | 0.927±0.005 | 0.927±0.006 |

Table 31 shows the results of CPC-CMS. ALBERT is the best model with and without considering the time factor. This indicates that the classification performance of ALBERT is excellent on the Case 3 dataset, and it even makes up for the drawback of high time consumption.

**Table 29.** The evaluation results on Case 3 testing dataset

| Model | Accuracy | Precision | Recall | F1-score | Specificity | MCC | Kappa |
|---|---|---|---|---|---|---|---|
| LSVC | 0.896 | 0.889 | 0.888 | 0.887 | 0.940 | 0.840 | 0.839 |
| Bernoulli Naive Bayes | 0.612 | 0.609 | 0.588 | 0.572 | 0.802 | 0.422 | 0.400 |
| Random Forest | 0.748 | 0.789 | 0.678 | 0.673 | 0.860 | 0.614 | 0.591 |
| Logistic Regression | 0.896 | 0.889 | 0.886 | 0.887 | 0.947 | 0.838 | 0.838 |
| XGBoost | 0.880 | 0.876 | 0.862 | 0.868 | 0.938 | 0.814 | 0.813 |
| LSTM | 0.895 | 0.897 | 0.895 | 0.896 | 0.948 | 0.838 | 0.838 |
| ALBERT | 0.947 | 0.947 | 0.947 | 0.947 | 0.973 | 0.918 | 0.918 |

**Table 30.** The efficiency result of the model on Case 3 dataset

| | LSVC | Bernoulli Naive Bayes | Random Forest | Logistic Regression | XGBoost | LSTM | ALBERT |
|---|---|---|---|---|---|---|---|
| Running time | 14.281s | 0.282s | 131.076s | 5.736s | 31.268s | 119.574s | 4004.657s |
| Efficiency | 0.997 | 1.000 | 0.967 | 0.999 | 0.992 | 0.970 | 0.000 |

**Table 31.** CPC-CMS results on Case 3 dataset

| Results without considering the efficiency factor | | | | | | | |
|---|---|---|---|---|---|---|---|
| | **LSVC** | **Bernoulli Naive Bayes** | **Random Forest** | **Logistic Regression** | **XGBoost** | **LSTM** | **ALBERT** |
| **Score** | 0.872 | 0.533 | 0.68 | 0.872 | 0.852 | 0.876 | 0.937 |
| **Rank** | 3 | 7 | 6 | 4 | 5 | 2 | 1 |
| **Results with considering the efficiency factor** | | | | | | | |
| | **LSVC** | **Bernoulli Naive Bayes** | **Random Forest** | **Logistic Regression** | **XGBoost** | **LSTM** | **ALBERT** |
| **Score** | 0.882 | 0.569 | 0.702 | 0.882 | 0.863 | 0.884 | 0.895 |
| **Rank** | 4 | 7 | 6 | 3 | 5 | 2 | 1 |

### 6.5. Discussions

After the simulations of the three cases are performed and analyzed, it can be concluded that LSTM and ALBERT are prone to overfitting for small- and medium-sized datasets with around 50,000 documents. It is necessary to optimize these models by adjusting parameters or introducing regularization, dropout, and an early stopping mechanism. All the models achieve stable performance on the validation set. In terms of performance on the test set, ALBERT ranks first based on the seven evaluation criteria in the three cases, but it consumes the longest time. We cannot conclude that ALBERT is the best model, since judging the superiority or inferiority of a model based on intuition or a single evaluation criterion is not rigorous, and it may lead to contradictions. Therefore, the CPC-CMS proposed in this study is used to obtain clear and fair comparison results.

The CPC-CMS results with and without the time factor for the three cases are shown in Table 32 and Table 33. respectively. ALBERT is the best model when not considering the time factor. If only high-quality classification results are pursued, ALBERT should be chosen without hesitation. However, if the time factor is considered, XGBoost has the highest score in Case 1, Random Forest has the highest score in Case 2, and ALBERT has the highest score in Case 3. Therefore, a balance must be struck between performance and time consumption when selecting the best model.

Table 32. The CPC-CMS results (without Efficiency) on three cases

| | LSVC | Bernoulli Naive Bayes | Random Forest | Logistic Regression | XGBoost | LSTM | ALBERT |
|---|---|---|---|---|---|---|---|
| **Score (Case 1)** | 0.754 | 0.607 | 0.657 | 0.741 | 0.788 | 0.718 | 0.811 |
| **Score (Case 2)** | 0.862 | 0.791 | 0.903 | 0.865 | 0.808 | 0.897 | 0.916 |
| **Score (Case 3)** | 0.872 | 0.533 | 0.68 | 0.872 | 0.852 | 0.875 | 0.937 |
| **Rank (Case 1)** | 3 | 7 | 6 | 4 | 2 | 5 | 1 |
| **Rank (Case 2)** | 4 | 7 | 2 | 5 | 6 | 3 | 1 |
| **Rank (Case 3)** | 4 | 7 | 6 | 3 | 5 | 2 | 1 |

**Table 33.** The CPC-CMS results (with Efficiency) on three cases

| | LSVC | Bernoulli Naive Bayes | Random Forest | Logistic Regression | XGBoost | LSTM | ALBERT |
|---|---|---|---|---|---|---|---|
| **Score (Case 1)** | 0.770 | 0.637 | 0.678 | 0.760 | 0.802 | 0.737 | 0.778 |
| **Score (Case 2)** | 0.873 | 0.807 | 0.909 | 0.876 | 0.823 | 0.887 | 0.876 |
| **Score (Case 3)** | 0.882 | 0.569 | 0.702 | 0.882 | 0.863 | 0.884 | 0.895 |
| **Rank (Case 1)** | 3 | 7 | 6 | 4 | 1 | 5 | 2 |
| **Rank (Case 2)** | 5 | 7 | 1 | 3 | 6 | 2 | 3 |
| **Rank (Case 3)** | 3 | 7 | 6 | 3 | 5 | 2 | 1 |

## 7. Comparisons of MCDM methods

PCNP [9] [10] [11] is proposed as an alternative to the AHP [12] [13], considering a paired interval scale, instead of a paired ratio scale, to be more reasonable. CPC is a core technique of PCNP. The weighted average on a weighted decision matrix is the aggregation and ranking method for both standard PCNP and AHP. The Technique for Order of Preference by Similarity to Ideal Solution (TOPSIS) [42] and Multi-Objective Optimization by Ratio Analysis (MOORA) [43] may be alternatives to the well-known weighted average, and are chosen for comparison, although no research has verified which aggregation and ranking method is superior to the others. The pyDecision Python package [44] is used to implement TOPSIS and MOORA. PCNP and AHP are implemented in this study. Aggregation and ranking methods requiring extra parameters, such as Preference Ranking Organization METHod for Enrichment of Evaluations (PROMETHEE) [45] and ELimination Et Choix Traduisant la REalite (ELECTRE) [46], are not considered in this study, due to the many possible ways to specify the parameters that lead to diverse and uncertain results.

The CPC weight sets derived from the input of two POMs for comparative evaluation criteria without and with efficiency are presented in Table 7 and Table 8, respectively. To convert CPC weights into AHP weights, POMs of CPC are converted into Pairwise Reciprocal Matrices (PRMs) of the AHP using the numerical conversion schema from the paired interval scale {-8,…,,-1,0,1,…,8} of CPC to the paired ratio scale {1/9,…,1/2,1,2,…,9} of the AHP to represent the comparative linguistic labels {"Extremely less important than", …, "Weakly less important than", "Equal to", "Weakly more important than", …, "Extremely more important than"} [10] [11] . As in the example mentioned in Section 5.2, to interpret the statement that Precision is weakly more important than Accuracy, the mathematical form $v_{Accuracy} - v_{Precision} = -2$ in CPC will be converted to $v_{Accuracy}/v_{Precision} = 1/3$ in AHP. After conversion into two PRMs, the Consistency Ratios for PRMs with and without efficiency factors are 0.07 and 0.05, respectively, which are within the recommended range. Table 34 presents the weights based on CPC and AHP for two sets of criteria either including or excluding efficiency. Both methods produce the same ranks but different weights. AHP overestimates the highest value and underestimates the lowest value due to the inappropriate definitions of the paired ratio scale [10] [11].

**Table 34.** Weights based on CPC and AHP for two sets of criteria of including or excluding efficiency

| | | Accuracy | Precision | Recall | F1-score | Specificity | MCC | Kappa | efficiency |
|---|---|---|---|---|---|---|---|---|---|
| **Seven factors** | **CPC** | 0.079 | 0.115 | 0.13 | 0.161 | 0.087 | 0.237 | 0.191 | - |
| | **AHP** | 0.024 | 0.050 | 0.071 | 0.131 | 0.029 | 0.468 | 0.227 | - |
| **Eight factors** | **CPC** | 0.090 | 0.117 | 0.133 | 0.150 | 0.104 | 0.193 | 0.166 | 0.047 |
| | **AHP** | 0.036 | 0.066 | 0.098 | 0.146 | 0.047 | 0.380 | 0.21 | 0.017 |
| | **Rank** | 7 | 5 | 4 | 3 | 6 | 1 | 2 | 8 |

For the weighted decision matrices for all cases, the weights for CPC and AHP are presented in Table 34, evaluation values for Case 1 are presented in Table 15 and Table **16**, evaluation values for Case 2 are shown in Table 22 and Table 23, and evaluation values for Case 3 are presented in Tables 29 – 30. With respect to aggregation and ranking methods based on CPC, AHP, CPC-TOPSIS, AHP-TOPSIS, CPC-MOORA, and AHP-MOORA, Table 35–Table **37** present aggregated decision values and ranks without considering the efficiency factor for each case, respectively. Like the CPC weighting and ranking method discussed in the previous section, if time factor is excluded, ALBERT is the best for the three datasets. All ranking methods produce the same rank for Cases 1 and 2, while AHP-TOPSIS in Case 3 produces a slightly different rank from the others.

Considering the efficiency factor, Table 38 - 40 present the values and ranks based on different aggregation and ranking methods for each case, respectively. Like the CPC weighting and ranking method discussed in the previous section, if the time factor is included, no single classification model always performs better than the other models. ALBERT takes a very long time for training, leading to the time difference among the other methods being less significant. For Case 1, CPC, CPC-TOPSIS, CPC-MOORA, and AHP-TOPSIS support XGBoost as the best choice, while AHP and AHP-MOORA still support ALBERT. For Case 2, all aggregation and ranking methods support Random Forest. For Case 3, all methods support ALBERT, except for CPC-TOPSIS, which supports LSTM.

**Table 35.** Aggregated values and rank results without efficiency in case 1

| Value:(Rank) | LSVC | Bernoulli Naive Bayes | Random Forest | Logistic Regression | XGBoost | LSTM | ALBERT |
|---|---|---|---|---|---|---|---|
| CPC | 0.754:(3) | 0.607:(7) | 0.657:(6) | 0.741:(4) | 0.788:(2) | 0.718:(5) | 0.811:(1) |
| AHP | 0.732:(3) | 0.569:(7) | 0.622:(6) | 0.714:(4) | 0.765:(2) | 0.685:(5) | 0.788:(1) |
| CPC-TOPSIS | 0.734:(3) | 0.09:(7) | 0.299:(6) | 0.668:(4) | 0.879:(2) | 0.55:(5) | 0.977:(1) |
| AHP-TOPSIS | 0.772:(3) | 0.037:(7) | 0.287:(6) | 0.672:(4) | 0.907:(2) | 0.521:(5) | 0.992:(1) |
| CPC-MOORA | 0.391:(3) | 0.313:(7) | 0.339:(6) | 0.384:(4) | 0.41:(2) | 0.372:(5) | 0.422:(1) |
| AHP-MOORA | 0.396:(3) | 0.306:(7) | 0.335:(6) | 0.385:(4) | 0.413:(2) | 0.369:(5) | 0.425:(1) |

**Table 36.** Aggregated values and rank results without efficiency in case 2

| Value:(Rank) | LSVC | Bernoulli Naive Bayes | Random Forest | Logistic Regression | XGBoost | LSTM | ALBERT |
|---|---|---|---|---|---|---|---|
| CPC | 0.862:(5) | 0.791:(7) | 0.903:(2) | 0.865:(4) | 0.808:(6) | 0.879:(3) | 0.917:(1) |
| AHP | 0.843:(5) | 0.762:(7) | 0.89:(2) | 0.847:(4) | 0.782:(6) | 0.862:(3) | 0.905:(1) |
| CPC-TOPSIS | 0.566:(5) | 0.0:(7) | 0.893:(2) | 0.595:(4) | 0.141:(6) | 0.701:(3) | 1.0:(1) |
| AHP-TOPSIS | 0.562:(5) | 0.0:(7) | 0.893:(2) | 0.593:(4) | 0.135:(6) | 0.698:(3) | 1.0:(1) |
| CPC-MOORA | 0.378:(5) | 0.346:(7) | 0.396:(2) | 0.38:(4) | 0.354:(6) | 0.386:(3) | 0.402:(1) |
| AHP-MOORA | 0.378:(5) | 0.341:(7) | 0.399:(2) | 0.38:(4) | 0.35:(6) | 0.387:(3) | 0.406:(1) |

**Table 37.** Aggregated values and rank results without efficiency in case 3

| Value:(Rank) | LSVC | Bernoulli Naive Bayes | Random Forest | Logistic Regression | XGBoost | LSTM | ALBERT |
|---|---|---|---|---|---|---|---|
| CPC | 0.872:(3) | 0.533:(7) | 0.68:(6) | 0.872:(4) | 0.852:(5) | 0.876:(2) | 0.937:(1) |
| AHP | 0.856:(3) | 0.473:(7) | 0.64:(6) | 0.855:(4) | 0.833:(5) | 0.857:(2) | 0.928:(1) |
| CPC-TOPSIS | 0.843:(3) | 0:(7) | 0.369:(6) | 0.84:(4) | 0.791:(5) | 0.845:(2) | 1.0:(1) |
| AHP-TOPSIS | 0.843:(2) | 0:(7) | 0.38:(6) | 0.84:(4) | 0.791:(5) | 0.841:(3) | 1.0:(1) |
| CPC-MOORA | 0.405:(3) | 0.245:(7) | 0.314:(6) | 0.405:(4) | 0.395:(5) | 0.406:(2) | 0.435:(1) |
| AHP-MOORA | 0.409:(3) | 0.224:(7) | 0.305:(6) | 0.408:(4) | 0.397:(5) | 0.409:(2) | 0.443:(1) |

**Table 38.** Aggregated values and rank results with efficiency in case 1

| Value:(Rank) | LSVC | Bernoulli Naive Bayes | Random Forest | Logistic Regression | XGBoost | LSTM | ALBERT |
|---|---|---|---|---|---|---|---|
| CPC | 0.770:(3) | 0.637:(7) | 0.678:(6) | 0.76:(4) | 0.802:(1) | 0.737:(5) | 0.778:(2) |
| AHP | 0.743:(3) | 0.588:(7) | 0.636:(6) | 0.727:(4) | 0.776:(2) | 0.7:(5) | 0.782:(1) |
| CPC-TOPSIS | 0.754:(2) | 0.323:(7) | 0.399:(6) | 0.705:(3) | 0.882:(1) | 0.611:(5) | 0.691:(4) |
| AHP-TOPSIS | 0.765:(3) | 0.12:(7) | 0.294:(6) | 0.675:(4) | 0.899:(1) | 0.535:(5) | 0.892:(2) |
| CPC-MOORA | 0.392:(3) | 0.32:(7) | 0.342:(6) | 0.386:(4) | 0.408:(1) | 0.374:(5) | 0.401:(2) |
| AHP-MOORA | 0.395:(3) | 0.31:(7) | 0.336:(6) | 0.386:(4) | 0.412:(2) | 0.371:(5) | 0.417:(1) |

**Table 39.** Aggregated values and rank results with efficiency in case 2

| Value:(Rank) | LSVC | Bernoulli Naive Bayes | Random Forest | Logistic Regression | XGBoost | LSTM | ALBERT |
|---|---|---|---|---|---|---|---|
| CPC | 0.873:(5) | 0.807:(7) | 0.909:(1) | 0.876:(4) | 0.823:(6) | 0.887:(2) | 0.876:(3) |
| AHP | 0.852:(5) | 0.776:(7) | 0.896:(1) | 0.856:(4) | 0.794:(6) | 0.87:(3) | 0.894:(2) |
| CPC-TOPSIS | 0.705:(4) | 0.466:(7) | 0.919:(1) | 0.722:(3) | 0.506:(6) | 0.786:(2) | 0.534:(5) |
| AHP-TOPSIS | 0.579:(5) | 0.175:(7) | 0.896:(1) | 0.608:(4) | 0.225:(6) | 0.708:(3) | 0.825:(2) |
| CPC-MOORA | 0.38:(5) | 0.35:(7) | 0.396:(1) | 0.381:(4) | 0.358:(6) | 0.386:(2) | 0.383:(3) |
| AHP-MOORA | 0.379:(5) | 0.344:(7) | 0.398:(1) | 0.38:(4) | 0.353:(6) | 0.387:(3) | 0.398:(2) |

**Table 40.** Aggregated values and rank results with efficiency in case 3

| Value:(Rank) | LSVC | Bernoulli Naive Bayes | Random Forest | Logistic Regression | XGBoost | LSTM | ALBERT |
|---|---|---|---|---|---|---|---|
| CPC | 0.882:(4) | 0.569:(7) | 0.702:(6) | 0.882:(3) | 0.863:(5) | 0.884:(2) | 0.895:(1) |
| AHP | 0.864:(3) | 0.502:(7) | 0.658:(6) | 0.863:(4) | 0.842:(5) | 0.865:(2) | 0.915:(1) |
| CPC-TOPSIS | 0.848:(2) | 0.205:(7) | 0.411:(6) | 0.846:(3) | 0.798:(4) | 0.851:(1) | 0.795:(5) |
| AHP-TOPSIS | 0.844:(2) | 0.059:(7) | 0.379:(6) | 0.84:(4) | 0.792:(5) | 0.842:(3) | 0.941:(1) |
| CPC-MOORA | 0.404:(3) | 0.257:(7) | 0.32:(6) | 0.404:(4) | 0.396:(5) | 0.405:(2) | 0.413:(1) |
| AHP-MOORA | 0.408:(3) | 0.235:(7) | 0.309:(6) | 0.407:(4) | 0.397:(5) | 0.408:(2) | 0.433:(1) |

## 8. Sensitivity Analysis of CPC-CMS

To evaluate the robustness of the CPC-CMS framework and ensure that the model evaluation results are not biased by the baseline reference weights, a sensitivity analysis is conducted across seven distinct weight configurations for each of the three case studies (21 scenario evaluations in total). Sensitivity analysis assesses how variations in the criteria weights impact the final ranking of the classification models, ensuring that the selected model is not overly dependent on slight fluctuations in subjective human judgment.

### 8.1. Perturbation Methodology

The sensitivity analysis systematically perturbs the weight of a target criterion and observes the resulting changes in the final model rankings. For this experimental setup, the "Efficiency" criterion is selected as the primary target for perturbation. The baseline weights are obtained from Table 8.

The baseline weight of the target criterion is adjusted by a perturbation level, $p$, which ranges from -30% to 30% in 10% increments. The newly perturbed target weight is calculated as:

$$w'_{target} = w_{target} \times (1 + p). \tag{47}$$

To ensure the sum of all criteria weights remains exactly 1, the weights of the remaining non-target criteria are proportionally redistributed. For any non-target criterion $i$, the new redistributed weight $w'_i$ is calculated using the following formula:

$$w'_i = \frac{w_i}{\sum w_{others}} \times \left(1 - w'_{target}\right) \tag{48}$$

**Table 41.** Perturbed Criteria Weights for Sensitivity Analysis

| Perturbation | Accuracy | Precision | Recall | F1 | Specificity | MCC | Kappa | Efficiency |
|---|---|---|---|---|---|---|---|---|
| -30% | 0.091 | 0.119 | 0.135 | 0.152 | 0.106 | 0.196 | 0.168 | 0.033 |
| -20% | 0.091 | 0.118 | 0.134 | 0.151 | 0.105 | 0.195 | 0.168 | 0.038 |
| -10% | 0.09 | 0.118 | 0.134 | 0.151 | 0.105 | 0.194 | 0.167 | 0.042 |
| 0 | 0.09 | 0.117 | 0.133 | 0.15 | 0.104 | 0.193 | 0.166 | 0.047 |
| 10% | 0.09 | 0.116 | 0.132 | 0.149 | 0.103 | 0.192 | 0.165 | 0.052 |
| 20% | 0.089 | 0.116 | 0.132 | 0.149 | 0.103 | 0.191 | 0.164 | 0.056 |
| 30% | 0.089 | 0.115 | 0.131 | 0.148 | 0.102 | 0.19 | 0.164 | 0.061 |

The results of the perturbed criteria weights for the sensitivity analysis are illustrated in Table 41. Manipulating the Efficiency weight from its baseline of 0.047 dynamically alters the weights of all other evaluation criteria. For instance, when the Efficiency weight is increased by 30% to reach 0.061, the remaining criteria weights must proportionally decrease to maintain a total sum of 1.0, ensuring the validity of the weighted decision matrix across all perturbation levels.

### 8.2. Spearman's Rank Correlation Test

The weighted decision matrix is recalculated across the datasets of the three cases using the newly generated weight arrays in Table 41. This yields perturbed comprehensive scores and new rank orders for the seven baseline models. To quantify the stability of the model rankings under these weight perturbations, Spearman's rank correlation coefficient ($r_s$) is utilized as the primary statistical measure.

$$r_s = 1 - \frac{6\sum d_i^2}{n(n^2-1)} \quad (49)$$

The coefficient $r_s$ measures the statistical dependence between two ranked variables, assessing how well their relationship can be described using a monotonic function. Here, $d_i$ is the difference between the ranks of corresponding variables, and $n$ is the total number of observations. The algorithm compares the baseline ranks (at 0% perturbation) directly against the newly computed ranks at every varying perturbation level. If a calculated $r_s$ value is close to 1.0, the weight variation has a negligible effect on the overall model selection, demonstrating high stability. Conversely, a significant drop in $r_s$ would signal that the framework is highly sensitive to the specific weight assigned to the Efficiency criterion, which would warrant careful consideration when relying on the CPC-CMS framework for time-sensitive applications. As presented in Table 42. , the framework demonstrates high robustness, though sensitivity varies slightly depending on the specific case characteristics:

- **Case 1:** The rankings are incredibly stable, maintaining an $r_s$ of 1.0 from a -30% to a 10% perturbation level. Only at the extreme 20% and 30% perturbations does the correlation dip slightly to 0.964 (due to LSVC and ALBERT swapping second and third ranks), indicating negligible impact on the overall decision.

- **Case 2:** The model selection remains strictly stable up to the 0% baseline. At positive perturbations (+10% to +30%), $r_s$ drops to 0.893 as shifts occur among models like ALBERT and LSVC. However, the top-ranked model (Random Forest) remains entirely unaffected regardless of the perturbation.

**Table 42.** Model rankings and $r_s$ under varying efficiency weights across three cases

| Case | *P* | Efficiency Weight | $r_s$ | LSVC | BNB | RF | LR | XGBoost | LSTM | ALBERT |
|---|---|---|---|---|---|---|---|---|---|---|
| | -30% | 0.033 | 1 | 3 | 7 | 6 | 4 | 1 | 5 | 2 |
| | -20% | 0.038 | 1 | 3 | 7 | 6 | 4 | 1 | 5 | 2 |
| | -10% | 0.042 | 1 | 3 | 7 | 6 | 4 | 1 | 5 | 2 |
| 1 | 0% | 0.047 | 1 | 3 | 7 | 6 | 4 | 1 | 5 | 2 |
| | 10% | 0.052 | 1 | 3 | 7 | 6 | 4 | 1 | 5 | 2 |
| | 20% | 0.056 | 0.964 | 2 | 7 | 6 | 4 | 1 | 5 | 3 |
| | 30% | 0.061 | 0.964 | 2 | 7 | 6 | 4 | 1 | 5 | 3 |
| | -30% | 0.033 | 0.964 | 5 | 7 | 1 | 4 | 6 | 3 | 2 |
| | -20% | 0.038 | 1 | 5 | 7 | 1 | 4 | 6 | 2 | 3 |
| | -10% | 0.042 | 1 | 5 | 7 | 1 | 4 | 6 | 2 | 3 |
| 2 | 0% | 0.047 | 1 | 5 | 7 | 1 | 4 | 6 | 2 | 3 |
| | 10% | 0.052 | 0.893 | 4 | 7 | 1 | 3 | 6 | 2 | 5 |
| | 20% | 0.056 | 0.893 | 4 | 7 | 1 | 3 | 6 | 2 | 5 |
| | 30% | 0.061 | 0.893 | 4 | 7 | 1 | 3 | 6 | 2 | 5 |
| | -30% | 0.033 | 0.964 | 3 | 7 | 6 | 4 | 5 | 2 | 1 |
| | -20% | 0.038 | 0.964 | 3 | 7 | 6 | 4 | 5 | 2 | 1 |
| | -10% | 0.042 | 0.964 | 3 | 7 | 6 | 4 | 5 | 2 | 1 |
| 3 | 0% | 0.047 | 1 | 4 | 7 | 6 | 3 | 5 | 2 | 1 |
| | 10% | 0.052 | 1 | 4 | 7 | 6 | 3 | 5 | 2 | 1 |
| | 20% | 0.056 | 1 | 4 | 7 | 6 | 3 | 5 | 2 | 1 |
| | 30% | 0.061 | 0.786 | 3 | 7 | 6 | 2 | 5 | 1 | 4 |

- **Case 3:** The framework shows strong stability at minor variations but exhibits the most significant sensitivity at the maximum 30% perturbation, where the $r_s$ drops to 0.786. In this extreme scenario, ALBERT drops from first to fourth place, and LSTM takes the top rank, underscoring that for datasets sharing Case 3's profile, aggressive changes to efficiency weighting can substantially alter the optimal model choice.

Overall, the predominantly high $r_s$ scores (mostly remaining above 0.893) across the three cases validate that the CPC-CMS framework provides robust and reliable model rankings despite natural variations in human judgment regarding criteria importance.

## 9. Conclusions

This study proposes the Cognitive Pairwise Comparison Classification Model Selection (CPC-CMS) framework to achieve human-centered classification model selection for document-level sentiment analysis. By utilizing CPC based on the preferences of domain experts to generate weights for eight evaluation criteria, the framework identifies the optimal model via a weighted selection matrix. Empirical simulations across seven models in three distinct case studies indicate that ALBERT is the top-performing model when execution time is excluded. However, when both classification performance and time consumption are evaluated simultaneously, the results demonstrate that no single model universally outperforms the others, reflecting the uncertainty and diversity of the underlying dataset patterns.

These conclusions are strongly supported by comparative evaluations against other aggregation and ranking methods, including AHP, CPC-TOPSIS, AHP-TOPSIS, CPC-MOORA, and AHP-MOORA. Furthermore, a sensitivity analysis using Spearman's Rank Correlation Test confirms the robustness of the proposed CPC-CMS framework. Notably, one of the case studies involves a mental health dataset, demonstrating the framework's practical utility for sentiment analysis and model selection within the medical and healthcare domains.

While the conceptual framework of the CPC-CMS approach is highly adaptable and can be generalized to other classification applications, the specific empirical model rankings derived in this study must be interpreted within their precise experimental boundaries. To avoid overgeneralization, it is important to note that the current findings are strictly conditional upon:

- Dataset Domain: The use of short, informal, and noisy English-language social media corpora;
- Criteria and Preference Weights: The specific selection of the eight evaluation metrics and preference matrices defined in our baseline and sensitivity scenarios;
- Computing Environment: The specific GPU/CPU hardware setup used for execution time benchmarks, as latency and memory metrics naturally vary across different architectures.

Future work will expand the evaluation of this framework across long-form documents, multilingual datasets, and edge-device hardware environments to thoroughly test its adaptability under broader operational constraints.

## Appendix

Summary of essential notations.

| Symbol | Description |
|---|---|
| $i, j, k$ | Local index variables |
| $B$ | Subjective judgment of the pairwise opposite matrix using pairwise interval scales |
| $\tilde{B}$ | Ideal pairwise matrix |
| $D$ | Document |

| $E$ | Sum of word embedding, position embedding, and sentence embedding |
|---|---|
| $E_{token}$ | Word embeddings |
| $E_{pos}$ | Position embeddings |
| $E_{seg}$ | Sentence embeddings |
| $G_i$ | Comprehensive score which is the weighted sum of the $i$-th model on all evaluation criteria |
| $M$ | Classification baseline model |
| $N$ | Total number of documents |
| $S$ | Results matrix of the $m$ models on the $n$ evaluation criteria |
| $S_{ij}$ | Result of the $i$-th model on the $j$-th evaluation criterion |
| $T$ | Time set of all models |
| $T_i$ | Running time of the $i$-th model |
| $V$ | Ideal utility set |
| $W$ | Weight set of all evaluation criteria |
| $b_{ij}$ | Comparison score of the difference between the priorities of two features |
| $m$ | Number of models |
| $n$ | Number of the evaluation criteria |
| $s_i$ | The $i$-th sentence |
| $t$ | a word |
| $v_i$ | The $i$-th ideal utility |
| $w_j$ | Weight of the $j$-th evaluation criterion |
| $x$ | Token sequence |
| $x_i$ | The $i$-th token |
| $y$ | A set of true categories |
| $\alpha$ | number of sentences |
| $\kappa$ | Average value of the feature weight in the paired assessment scale |
| $p_o$ | Proportion of the model's prediction being consistent with the true label |
| $p_e$ | Expected ratio where the predicted label is consistent with the true label |
| $TP$ | Number of positive samples successfully identified as positive |
| $FP$ | Number of negative samples wrongly classified as positive |
| $FN$ | Number of positive samples wrongly classified as negative |
| $TN$ | Number of negative samples successfully identified as negative |

**Funding**

This research has not been funded by any company or organization.

**Author Contributions**

- Jianfei LI: Conceptualization – CMS (50%), Methodology – CMS (80%), Literature Review (80%), Software – CMS, Simulation – ML, Validation (30%), Writing – Original Draft Preparation.
- Kevin Kam Fung YUEN: Conceptualization – CPC (50%), Methodology – CMS (20%), Methodology – MCDM, Literature Review (20%), Software – MCDM, Simulation – MCDM, Sensitivity Analysis, Validation (70%), Revisions, Writing – Review and Editing, Responses to Reviewers, Supervision.

**Conflict of interest**

The authors declare that they have no conflicts of interest regarding this work.

**Data availability**

The dataset is described in Section 4.1.

**Source code**

Essential code is presented in the paper. The source code will be posted on the corresponding author's GitHub repository (kkfyuen) or another outlet.

## References

[1] Liu, B., Zhang, L.: A Survey of Opinion Mining and Sentiment Analysis. In: Aggarwal, C., Zhai, C. (eds) Mining Text Data. Springer, Boston, MA. (2012). https://doi.org/10.1007/978-1-4614-3223-4_13

[2] Jain, P.K., Pamula, R., Srivastava, G.: A systematic literature review on machine learning applications for consumer sentiment analysis using online reviews. Computer Science Review, Volume 41, (2021). https://doi.org/10.1016/j.cosrev.2021.100413

[3] Rodríguez-Ibánez, M., Casánez-Ventura, A., Castejón-Mateos, F., Cuenca-Jiménez, P.-M.: A review on sentiment analysis from social media platforms. Expert Systems with Applications, Volume 223, (2023). https://doi.org/10.1016/j.eswa.2023.119862.

[4] Baker, M.R., Mohammed, E.Z., Jihad, K.H.: Prediction of Colon Cancer Related Tweets Using Deep Learning Models. In: Abraham, A., Pllana, S., Casalino, G., Ma, K., Bajaj, A. (eds) Intelligent Systems Design and Applications. ISDA 2022. Lecture Notes in Networks and Systems. Volume 646. Springer, Cham. (2023). https://doi.org/10.1007/978-3-031-27440-4_50

[5] Rognone, L., Hyde, S., Zhang, S.S.: News sentiment in the cryptocurrency market: An empirical comparison with Forex. International Review of Financial Analysis, Volume 69, (2020). https://doi.org/10.1016/j.irfa.2020.101462

[6] Behdenna, S., Barigou, F., Belalem, G.: Sentiment Analysis at Document Level. In: Unal, A., Nayak, M., Mishra, D.K., Singh, D., Joshi, A. (eds) Smart Trends in Information Technology and Computer Communications. SmartCom 2016. Communications in Computer and Information Science, Volume 628. Springer, Singapore. (2016). https://doi.org/10.1007/978-981-10-3433-6_20

[7] Mao, Y.Y., Liu, Q., Zhang, Y.: Sentiment analysis methods, applications, and challenges: A systematic literature review. Journal of King Saud University - Computer and Information Sciences, Volume 36, Issue 4, (2024). https://doi.org/10.1016/j.jksuci.2024.102048.

[8] Raschka, S.: Model evaluation, model selection, and algorithm selection in machine learning, arXiv preprint, 2018, https://doi.org/10.48550/arXiv.1811.12808.

[9] Yuen, K.K.F.: Cognitive pairwise comparison forward feature selection with deep learning for astronomical object classification with sloan digital sky survey. Discov Artif Intell 4, 39 (2024). https://doi.org/10.1007/s44163-024-00140-5

[10] Yuen, K.K.F.: Cognitive network process with fuzzy soft computing technique in collective decision aiding. The Hong Kong Polytechnic University, Ph.D. thesis. (2009).

[11] Yuen, K.K.F.: The pairwise opposite matrix and its cognitive prioritization operators: The ideal alternatives of the pairwise reciprocal matrix and analytic prioritization operators. J. Oper. Res. Soc., vol. 63, pp. 322-338, (2012).

[12] T.L. Saaty, "A scaling method for priorities in hierarchical structures," *Journal of Mathematical Psychology,* vol. 15, pp. 234-281, 1977.

[13]T. L. Saaty, Analytic Hierarchy Process: Planning, Priority, Setting, Resource Allocation, New York: McGraw-Hill, 1980.

[14] Zhao, J., Liu, K., Xu, L.H.: Sentiment Analysis: Mining Opinions, Sentiments, and Emotions. Computational Linguistics, 42 (3), 595-598, (2016). https://doi.org/10.1162/COLI_r_00259

[15] Rhanoui, M., Mikram, M., Yousfi, S., Barzali, S.: A CNN-BiLSTM Model for Document-Level Sentiment Analysis. Machine Learning and Knowledge Extraction, 1 (3), 832-847, (2019). https://doi.org/10.3390/make1030048

[16] Pu, X., Wu, G., Yuan, C.: Exploring overall opinions for document level sentiment classification with structural SVM. Multimedia Systems 25, 21–33, (2019). https://doi.org/10.1007/s00530-017-0550-0

[17] Zhai, S., Zhang, Z.: Semisupervised Autoencoder for Sentiment Analysis. Proceedings of the AAAI Conference on Artificial Intelligence, 30(1), (2016). https://doi.org/10.1609/aaai.v30i1.10159

[18] Xu, J., Chen, D., Qiu, X., Huang, X.: Cached Long Short-Term Memory Neural Networks for Document-Level Sentiment Classification. https://doi.org/10.48550/arXiv.1610.04989 (2016). Accessed 26 June 2025

[19] Sentiment Analysis for Mental Health. https://www.kaggle.com/datasets/suchintikasarkar/sentiment-analysis-for-mental-health/data (2024). Accessed 26 June 2025

[20] Salton, G., Buckley, C.: Term-weighting approaches in automatic text retrieval. Information Processing & Management, Volume 24, Issue 5, Pages 513-523, (1988). https://doi.org/10.1016/0306-4573(88)90021-0

[21] Mikolov, T., Sutskever, I., Chen, K., Corrado, G., Dean, J.: Distributed representations of words and phrases and their compositionality. https://doi.org/10.48550/arXiv.1310.4546 (2013). Accessed 26 June 2025

[22] Devlin, J., Chang, M.-W., Lee, K., Toutanova, K.: BERT: Pre-training of Deep Bidirectional Transformers for Language Understanding. https://doi.org/10.48550/arXiv.1810.04805 (2019). Accessed 26 June 2025

[23] Lan, Z., Chen, M., Goodman, S., Gimpel, K., Sharma, P., Soricut, R.: ALBERT: A Lite BERT for Self-supervised Learning of Language Representations.https://doi.org/10.48550/arXiv.1909.11942 (2020).

Accessed 26 June 2025

[24] Lewis, D.D.: Naive (Bayes) at forty: The independence assumption in information retrieval. In: Nédellec, C., Rouveirol, C. (eds) Machine Learning: ECML-98. ECML 1998. Lecture Notes in Computer Science, vol 1398. Springer, Berlin, Heidelberg. (1998). https://doi.org/10.1007/BFb0026666

[25] Breiman, L.: Random Forests. Machine Learning 45, 5–32, (2001). https://doi.org/10.1023/A:1010933404324

[26] Sohil, F., Sohali, M.U., Shabbir, J.: An introduction to statistical learning with applications in R. In: James, G., Witten, D., Hastie, T., Tibshirani, R. (eds) New York, Springer Science and Business Media, 2013, eISBN: 978-1-4614-7137-7. Statistical Theory and Related Fields, 6 (1), 87, (2021). https://doi.org/10.1080/24754269.2021.1980261

[27] Fan, R.-E., Chang, K.-W., Hsieh, C.-J., Wang, X.-R., Lin, C.-J.: LIBLINEAR: A Library for Large Linear Classification. The Journal of Machine Learning Research, 9 (6/1/2008), 1871–1874, (2008). https://dl.acm.org/doi/10.5555/1390681.1442794

[28] Chen, T., Guestrin, C.: XGBoost: A Scalable Tree Boosting System. In Proceedings of the 22nd ACM SIGKDD International Conference on Knowledge Discovery and Data Mining (KDD '16). Association for Computing Machinery, New York, NY, USA, 785–794, (2016). https://doi.org/10.1145/2939672.2939785

[29] Hochreiter, S., Schmidhuber, J.: Long Short-Term Memory. Neural Computation, 9 (8): 1735–1780, (1997). https://doi.org/10.1162/neco.1997.9.8.1735

[30] Vaswani, A., Shazeer, N., Parmar, N., Uszkoreit, J., Jones, L., Gomez, A.N., Kaiser, L., Polosukhin, I.: Attention is all you need. https://doi.org/10.48550/arXiv.1706.03762 (2023). Accessed 26 June 2025

[31] Youden, W.J.: Index for rating diagnostic tests. Cancer, 3: 32-35, (1950). https://doi.org/10.1002/1097-0142(1950)3:1<32::aid-cncr2820030106>3.0.co;2-3

[32] Matthews, B.W.: Comparison of the predicted and observed secondary structure of T4 phage lysozyme. Biochimica et Biophysica Acta (BBA) - Protein Structure, Volume 405, Issue 2, Pages 442-451, (1975). https://doi.org/10.1016/0005-2795(75)90109-9

[33] Cohen, J.: A Coefficient of Agreement for Nominal Scales. Educational and Psychological Measurement, *20*(1), 37-46, (1960). https://doi.org/10.1177/001316446002000104

[34] Chicco, D., Jurman, G.: The advantages of the Matthews correlation coefficient (MCC) over F1 score and accuracy in binary classification evaluation. BMC Genomics 21, 6, (2020). https://doi.org/10.1186/s12864-019-6413-7

[35] Sokolova, M., Lapalme, G.: A systematic analysis of performance measures for classification tasks. Information Processing & Management, Volume 45, Issue 4, Pages 427-437, (2009). https://doi.org/10.1016/j.ipm.2009.03.002

[36] Scikit-learn. https://scikit-learn.org/. Accessed 26 June 2025

[37] Tensorflow. https://www.tensorflow.org/. Accessed 26 June 2025

[38] Transformers. https://github.com/huggingface/transformers. Accessed 26 June 2025

[39]Twitter Sentiment Analysis. https://www.kaggle.com/datasets/jp797498e/twitter-entity-sentiment-analysis (2021). Accessed 26 June 2025

[40] Gowda, C., Anirudh, Pai, A., Kumar A, C.: Twitter and Reddit Sentimental analysis Dataset [Data set]. Kaggle. https://doi.org/10.34740/KAGGLE/DS/429085 (2019). Accessed 26 June 2025

[41] nltk. https://www.nltk.org/. Accessed 26 June 2025

[42] C.-L. Hwang and K. Yoon, "Methods for multiple attribute decision making," in Multiple attribute decision making: methods and applications a state-of-the-art survey, Berlin, Springer Berlin Heidelberg, 1980, pp. 58-191.

[43] W. K. Brauers, Optimization Methods for a Stakeholder Society: A Revolution in Economic Thinking by Multi-objective Optimization, Springer : Springer , https://doi.org/10.1007/978-1-4419-9178-2, 2004.

[44] V. PEREIRA, M. BASILIO and C. SANTOS, "Enhancing Decision Analysis with a Large Language Model: pyDecision a Comprehensive Library of MCDA Methods in Python," Journal of Modelling in Managemen, vol. 21, no. 2, pp. 481–521,doi: https://doi.org/10.1108/JM2-04-2024-0118, 2026.

[45] B. Jean-Pierre, "L'ingéniērie de la Décision; Elaboration D'instruments D'aide à la Decision. La Méthode ProMEThEE," in L'aidea la Décision: Nature, Instruments et Perspectives d'Avenir, 1982.

[46] B. Roy, "Classement et choix en présence de points de vue multiples," Revue française d'informatique et de recherche opérationnelle, vol. 2, no. 8, pp. 57-75, 1968.